\documentclass[11pt]{article}
\usepackage{graphicx,fullpage,xcolor,comment,bm} 
\usepackage{amsmath,amsfonts, mathtools, amsthm, amssymb}
\usepackage[utf8]{inputenc}
\usepackage[english]{babel}
\usepackage[autostyle]{csquotes}
\usepackage{caption}
\usepackage{enumitem}
\usepackage[hidelinks]{hyperref}
\usepackage{adjustbox}
\usepackage[capitalize]{cleveref}
\usepackage[margin=1in]{geometry}
\usepackage{url}
\usepackage[
backend=biber,
style= alphabetic,
sorting=nty
]{biblatex}
\usepackage{blindtext}
\usepackage{multirow}
\usepackage{hhline}
\usepackage{authblk}

\title{\textbf{Vision-Language Models for Medical Report Generation and Visual Question Answering: A Review}}

\date{\vspace{-5ex}}

\author[1]{Iryna Hartsock}
\author[1]{Ghulam Rasool}
\affil[1]{Department of Machine Learning, H. Lee Moffitt Cancer Center \& Research Institute}

\begin{document}

\maketitle

\begin{abstract}
Medical vision-language models (VLMs) combine computer vision (CV) and natural language processing (NLP) to analyze visual and textual medical data. Our paper reviews recent advancements in developing VLMs specialized for healthcare, focusing on models designed for medical report generation and visual question answering (VQA). We provide background on NLP and CV, explaining how techniques from both fields are integrated into
VLMs to enable learning from multimodal data. Key areas we address include the exploration of medical vision-language datasets, in-depth analyses of architectures and pre-training strategies employed in recent noteworthy medical VLMs, and comprehensive discussion on evaluation metrics for assessing VLMs' performance in medical report generation and VQA. We also highlight current challenges and propose future directions, including enhancing clinical
validity and addressing patient privacy concerns. Overall, our review summarizes recent
progress in developing VLMs to harness multimodal medical data for improved healthcare
applications.
\end{abstract}

\section{Introduction}

The last decade has witnessed enormous progress in artificial intelligence (AI) and machine learning (ML), including the development of foundation models (FMs), large language models (LLMs), and vision-language models (VLMs). These AI/ML developments have started transforming several aspects of our daily lives, including healthcare. AI/ML can potentially transform the whole healthcare continuum by significantly optimizing and improving disease screening and diagnostic procedures, treatment planning, and post-treatment surveillance and care \cite{bajwa2021aihealthcare}. Various computer vision (CV) and natural language processing (NLP) models, more recently LLMs, have been instrumental in driving this transformative trend \cite{he2023llmsurvey, zhou2023llmsurvey2}. CV models have been trained and validated for various screening and diagnosis use cases leveraging radiology data from X-rays, mammograms, magnetic resonance imaging (MRI), computed tomography (CT), and others. Recently, AI models focused on digital pathology using histopathology and immunohistochemistry data have also shown significant advances in accurate disease diagnosis, prognosis, and biomarker identification \cite{WAQAS2023100255}. On the other hand, by training models using large datasets of medical literature, clinical notes, and other healthcare-related text, LLMs can extract insights from electronic health records (EHR) efficiently, assist healthcare professionals in generating concise summary reports, and facilitate the interpretation of patient information. Noteworthy examples of such LLMs include \emph{GatorTron} \cite{yang2022gatortron}, \emph{ChatDoctor} \cite{li2023chatdoctor}, \emph{Med-PaLM} (Medical Pathways Language Model) \cite{singhal2023medpalm} and \emph{Med-Alpaca} \cite{han2023medalpaca}. 

The healthcare data is inherently multimodal, and consequently, the AI/ML models often need to be trained using multiple data modalities, including text (e.g., clinical notes, radiology reports, surgical pathology reports, etc.), imaging (e.g., radiological scans, digitized histopathology slides, etc.), and tabular data (e.g., numerical data such as vitals or labs and categorical data such as race, gender, and others) \cite{acosta2022biomedicalai, shrestha2023medicalvlms, waqas2023multimodaloncology, tripathi2023multimodaloncologydataset, Mohsan2023}. In routine clinical practice, healthcare professionals utilize a combination of these data modalities for diagnosing and treating various conditions. Integrating information from diverse data modalities enhances the precision and thoroughness of disease assessments, diagnoses, treatment planning, and post-treatment surveillance. The need for AI/ML models to ingest, integrate, and learn from information stemming from varied data sources is the driving force for \emph{multimodal learning} \cite{huang2021multimod, waqas2023multimodaloncology}. 

The recent progress in multimodal learning has been driven by the development of \emph{vision-language models (VLMs)} \cite{gan2022vlmsurvey, chen2022vlp, Mohsan2023}. These cutting-edge models can analyze, interpret, and derive insights from both visual and textual data. In the medical domain, these models contribute to developing a more holistic understanding of patient information and improving the performance of ML models in various clinical tasks. Many of these models, like \emph{CLIP} (Contrastive Language–Image Pre-training) \cite{radford2021clip}, \emph{LLaVa} (Large Language and Vision Assistant) \cite{liu2023llava}, and \emph{Flamingo} \cite{alayrac2022flamingo} are tailored to healthcare domain through training on extensive medical datasets. Adapting VLMs for medical visual question-answering \cite{lin2023medvqa} is particularly noteworthy, empowering healthcare professionals to pose queries regarding medical images such as CT scans, MRIs, mammograms, ultrasound, X-rays, and more. The question-answering capability elevates the interactive nature of the AI/ML models in healthcare, facilitating dynamic and informative exchanges between healthcare providers and the AI system. Furthermore, adapting VLMs for medical report generation enables them to produce detailed and contextually relevant reports by amalgamating information from both visual and textual sources. This not only streamlines the documentation process but also ensures that the generated reports are comprehensive and accurately reflect the subtleties present in the data, further enhancing healthcare workflow efficiency.  

\begin{figure}[ht] 
\centering
\includegraphics[width=\linewidth]{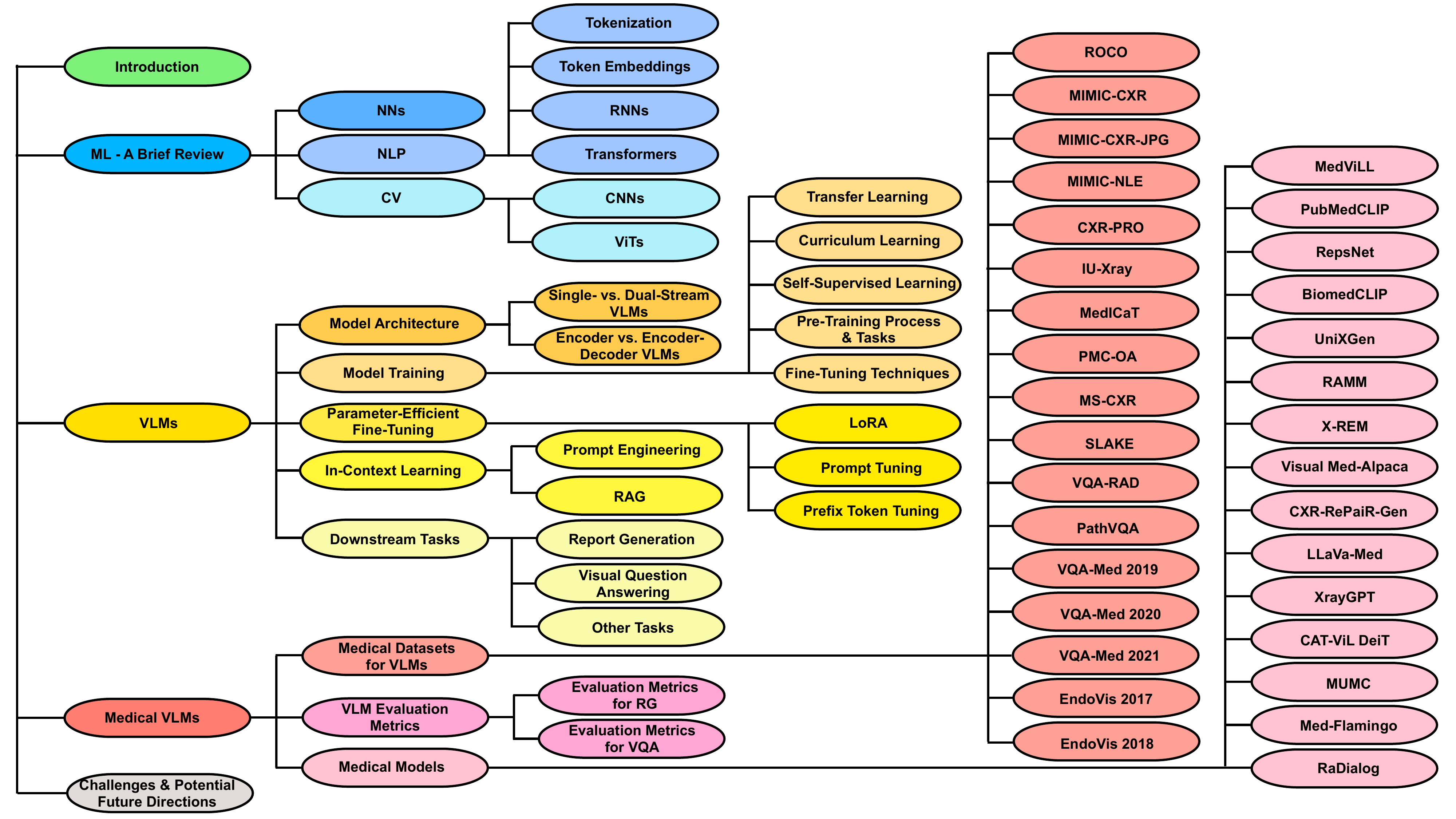}
\caption{Organization of the review paper.}
\label{fig:structure}
\end{figure}

In contrast to previous related surveys \cite{lin2023medvqa, ting2023mrgsurvey, shrestha2023medicalvlms}, this review focuses on the latest advancements in VLMs tailored for medical report generation and visual question-answering. 
The overall structure of this review is shown in \cref{fig:structure} and is outlined as follows. In \cref{sec:background}, we provide essential background on neural networks, CV, and NLP. In \cref{sec:vlm}, we delve into the exploration of VLMs' architectures, training strategies, and downstream tasks. The goal of \cref{sec:background} and \cref{sec:vlm}  is to ensure the accessibility of this review for readers, irrespective of their ML background. We split \cref{sec:medical_vlms} into three key sub-sections.
In \cref{subsec:datasets}, we describe 17 publicly available vision-language datasets. These datasets encompass medical image-text pairs or question-answer pairs related to medical images. Next, in \cref{subsec:metrics}, we meticulously outline the metrics and their formulas, where applicable, employed for evaluating VLMs in the context of report generation and visual question-answering tasks. 
In \cref{subsec:models}, we conduct a thorough review of 15 recent medical VLMs, with 14 of them being publicly available. To the best of our knowledge, most of these models have not been reviewed in any previous surveys. Finally, in \cref{sec:future_work}, we discuss the current challenges within the field of medical VLMs, offering insights into potential research directions that could profoundly influence their future development. The list of medical VLMs and datasets can also be found on \href{https://github.com/lab-rasool/Awesome-Medical-VLMs-and-Datasets/tree/main}{GitHub}.

\section{Machine Learning (ML) - A Brief Review}\label{sec:background}

\subsection{Neural Networks (NNs)}
ML and AI, as we understand them today, began to take shape in the late 1940s and early 1950s \cite{baldi2021dlscience}. NNs stand out as classical ML models, drawing inspiration from the structure and functioning of the human brain. They are composed of layers of interconnected nodes or neurons arranged into an input layer, an output layer, and multiple intermediate layers called hidden layers. The basic NN is a ``feedforward NN'', where neurons can be numbered in such a way that a connection from neuron $i$ to neuron $j$ can exist if and only if $i <j$ \cite{baldi2021dlscience}. In any NN, the connections between nodes carry weight, and neurons utilize ``activation functions'' on their inputs.  Activation functions play a crucial role by introducing non-linearity to the model, enabling it to learn complex nonlinear mappings between inputs and outputs. Common activation functions include the \emph{sigmoid}, \emph{hyperbolic tangent (tanh)}, and \emph{Rectified Linear Unit (ReLU)}. NNs utilize a loss function to quantify the difference between the predicted outputs and the actual targets. The loss function produces a scalar value, and the goal during training is to minimize this loss value. 

\emph{Backpropagation}, short for \emph{backward propagation of errors}, is a key algorithm for training deep neural networks. During the forward pass, input data is fed through the network, predictions are generated, and a scalar loss value is calculated using the loss function. During backpropagation, we calculate the gradient of the loss function with respect to the weights of the network. This gradient information is then used to update the weights in an effort to minimize the difference between predicted and target values. Backpropagation is an application of the chain rule for computing derivatives \cite{baldi2021dlscience}. After the backward pass, the optimization algorithm takes these gradients and adjusts the learnable parameters (weights and biases) of the NN, which, in turn, will result in the minimization of the loss value in the next batch. Common optimization methods include gradient descent, stochastic gradient descent (SGD), \cite{robbins1951sgd}, Adam (Adaptive Moment Estimation \cite{kingma2014adam}, and many others.

\subsection{Natural Language Processing (NLP)}

NLP is the analysis of linguistic data, most commonly in the form of textual data such as documents or publications, using computational methods \cite{verspoor2013nlp}. NLP encompasses a variety of tasks aimed at understanding, processing, and generating human language. \emph{Named entity recognition (NER)}, a prominent NLP task, focuses on identifying and classifying entities within the text, such as names of individuals, medical conditions, etc. For instance, in medical literature, NER can assist in extracting crucial information from documents. \emph{Text summarization} in NLP is widely used for generating coherent summaries of lengthy texts. \emph{Sentiment analysis} is a task that determines the emotional tone expressed in a given text, providing valuable insights for applications like social media monitoring or customer feedback analysis. \emph{Machine translation} is a fundamental NLP task in breaking down language barriers by automatically translating text from one language to another. \emph{Question answering} in NLP is directed at comprehending and responding to user queries, propelling advancements in virtual assistants and information retrieval.

\subsubsection{Tokenization}The first step in NLP is tokenization, which is the mechanism of splitting or fragmenting the sentences and words to their possible smallest morpheme, called a token. A morpheme is the smallest possible word after which it cannot be broken further \cite{rai2021token}. One example of a word-level tokenization method is \emph{whitespace tokenization}, which segments text based on whitespace characters. In many NLP applications, subword tokenization methods are preferred due to their effectiveness in handling out-of-vocabulary words.   \emph{WordPiece} \cite{wu2016wordpiece} begins by treating each character as a token, creating an initial vocabulary. Employing a flexible and adaptive merging strategy, WordPiece considers any pair of adjacent characters or subword units that enhance the overall likelihood of the training data. This likelihood reflects the model's probability of accurately representing the training data given its current state.
In contrast, \emph{Byte-Pair Encoding (BPE)} \cite{sennrich2016bpe} shares similarities with WordPiece but adheres to a more deterministic merging strategy. In each iteration, BPE merges the most frequent pair of adjacent characters or subword units, progressing toward a predefined vocabulary size.  \emph{Byte-level BPE} \cite{wang2020bbpe} operates at an even finer granularity, considering individual bytes rather than characters. Byte-level BPE extends the concept of subword tokenization to bytes, allowing it to capture more nuanced patterns at the byte level.

\subsubsection{Token Embeddings} Tokens are then often transformed into numerical vectors that capture semantic relationships between tokens, which are referred to as word or token embeddings. \emph{Word2Vec} \cite{mikolov2013wordvec} is a widely used word embedding technique that uses two models: Skip-Gram \cite{mikolov2013wordvec} and Continuous Bag of Words (CBOW) \cite{mikolov2013cbow}. In skip-gram, the model predicts context words given a target word, capturing semantic associations. Conversely, CBOW predicts the target word based on its context, emphasizing syntactic structures. In both models, the ``context word'' refers to words within a specified window around the target word, and the ``target word'' is the word for which predictions are made. Word2Vec is computationally efficient, making it well-suited for large datasets and general-purpose applications. \emph{Global Vectors (GloVe)} \cite{pennington2014glove}  is a word embedding model that distinguishes itself by capturing global semantic relationships. It focuses on the entire corpus rather than local context windows. The model builds a co-occurrence matrix based on the global statistics of word pairs and then employs an objective function to generate word vectors that reflect the ratio of co-occurrence probabilities. GloVe uses an implicit skip-gram approach, capturing word relationships globally, making it ideal for tasks that demand a holistic understanding of word connections. \emph{FastText} \cite{bojanowski2017fasttext} is another word embedding that is particularly effective for handling out-of-vocabulary words and morphologically rich languages. It adopts a sub-word approach, breaking words into n-grams, and employs a skip-gram training method similar to Word2Vec \cite{mikolov2013wordvec} to learn embeddings for these sub-word units. There are also word embeddings tailored to represent biomedical and clinical terms better. In tasks where the order of words is not essential, other feature extraction techniques can be effective, for example, \emph{bag-of-words (BoW)} in text classification or \emph{term frequency-inverse document frequency (tf-idf)} for information retrieval. 

In addition to general-purpose word embeddings, there are ones designed for biomedical and clinical terms.
\emph{BioWordVec} \cite{zhang2019biowordvec} incorporates MeSH (Medical Subject Headings) terms along with text from PubMed abstracts and employs the fastText \cite{bojanowski2017fasttext} algorithm to learn improved biomedical word embeddings. Another prominent approach is \emph{Cui2vec} \cite{beam2018cui2vec}, which leverages diverse multi-modal data derived from medical publications and clinical notes. Cui2vec systematically maps medical terms onto a common Concept Unique Identifier (CUI) space, followed by the construction of a co-occurrence matrix. This matrix captures instances where different CUIs appear together, which is a foundation for generating word embeddings using techniques such as GloVe \cite{pennington2014glove} or Word2Vec \cite{mikolov2013wordvec}. 
In most cases, it is common to add \emph{positional encodings} to capture the order of tokens in a sequence. Positional encoding vectors, often based on sinusoidal functions, systematically encode token positions, enriching embeddings with positional information for utilization in ML models tailored to specific NLP tasks \cite{ahmed2023transformers}.

\subsubsection{Recurrent Neural Networks (RNNs)} RNNs are widely employed for pattern detection in sequential data, encompassing diverse types such as genomic sequences, text, or numerical time series \cite{schmidt2019rnn}. Operating on the principle of preserving a form of memory, RNNs incorporate a cyclic structure by looping the output of a specific layer back to the input, facilitating the prediction of subsequent layer outputs. This mechanism empowers RNNs to adeptly model sequential and temporal dependencies, capturing information from preceding time steps within hidden states. Despite their capacity to retain information from past inputs, RNNs encounter challenges in preserving long-term dependencies within input sequences due to the vanishing gradient problem. To address this, several RNN variants, including Long Short-Term Memory (LSTM) \cite{hochreiter1997lstm} and Gated Recurrent Unit (GRU) \cite{cho2014gru} networks, have been designed to enhance their ability to capture and utilize long-range dependencies in sequential data.

\subsubsection{Transformers} 
In recent years, there has been a remarkable advancement in NLP mainly due to the development of the Transformer models \cite{vaswani2017attention}. Beyond incorporating embeddings and positional encodings, the Transformer architecture consists of an encoder that processes input data, represented by vectors obtained from embedded and positionally encoded tokens. The encoder-generated representation then serves as the input for the subsequent decoder, which transforms these vector representations into a relevant output tailored to the specific task at hand.  A defining characteristic of the Transformer lies in its \emph{self-attention mechanism}, notably the scaled dot-product attention, which proves instrumental in capturing intricate dependencies within sequences. This mechanism, employed in both the encoder and decoder, utilizes queries and keys during the attention process. Queries serve as projections of the input sequence, encapsulating information for attending to other positions, while keys represent the positions within the sequence. Enhanced by multi-head attention for parallelization, the self-attention mechanism enables the model to dynamically weigh different parts of the input sequence, fostering a nuanced understanding of contextual relationships. Each layer in both the encoder and decoder encompasses sub-layers, including a feedforward NN, further augmenting the model's capacity to capture intricate patterns within the data. In practice,  Transformers face limitations in effectively processing long sequences and exhibit less selectivity about relevant information when considering all positions in the sequence. Various techniques have been proposed to address these issues. One such approach, known as hierarchical attention \cite{yang2016hierarchical}, strategically reduces computational complexity and enhances contextual sensitivity by initially computing attention at the word level and then at the sentence level. Another notable advancement in attention algorithms is the FlashAttention \cite{dao2022flashattention} and FlashAttention-2 \cite{dao2023flashattention2}, designed to accelerate attention computations significantly.

The synergy between the enhanced computational power provided by Graphical Processing Units (GPUs) and the advancements in attention mechanisms has played a pivotal role in the development of large language models (LLMs). These models are meticulously trained on vast datasets with a very large number of parameters. Initial LLMs include but are not limited to, BERT (Bidirectional Encoder Representations from Transformers) \cite{devlin2019bert} (the largest version comprising $235$ M parameters),  ALBERT (A Lite BERT) \cite{lan2019albert} (with the largest variant at $12$ M parameters), and Megatron-LM \cite{shoeybi2020megatronlm} (the largest version featuring $1.2$ B parameters). The era of even larger LLMs began in $2020$, introducing models like  GPT-3 (the 3-generation Generative Pre-trained Transformer) \cite{brown2020gbt} ($175$ B parameters) and PaLM (Pathways Language Model) \cite{chowdhery2022palm} ($540$ B parameters). Some of the most recent LLMs are LLaMA (Large Language Model Meta AI) \cite{touvron2023llama}, Vicuna \cite{chiang2023vicuna}, Llama 2 \cite{touvron2023llama2}, and Mistral \cite{jiang2023mistral}. Note that encoder-only LLMs can be used to generate token embeddings (e.g.,  BERT \cite{devlin2019bert} or GatorTron \cite{yang2022gatortron}).

\subsection{Computer Vision (CV)}
CV involves interpreting and understanding the world from their images or videos \cite{ji2020cv}. Data in CV is encoded as numerical values representing the intensity or brightness of pixels. The extraction of visual patterns like edges, textures, and objects in images or video frames serves as building blocks for various CV tasks. \emph{Image classification} is the task of assigning a label to an entire image, determining the main object or scene. \emph{Object detection} involves identifying and locating multiple objects within an image, providing both labels and bounding boxes. \emph{Image segmentation} divides an image into meaningful segments, assigning a label to each pixel and outlining the boundaries of distinct objects or regions. Various ML techniques and models are utilized for these tasks \cite{mahadevka2022MLinCV}. 

\subsubsection{Convolutional Neural Networks (CNNs)}
CNNs represent a significant advancement in CV \cite{yamashita2018cnn}. Besides pooling and fully connected layers, CNNs also have convolution layers, which apply convolution operations to input data. During a convolution operation, a small filter or kernel slides over the input data. At each position, the filter performs element-wise multiplications with the local regions of the input. The results of these multiplications are then summed, creating a new value in the output feature map. This process is repeated across the entire input, capturing patterns and features at different spatial locations. The well-known CNNs include Residual Network (ResNet) \cite{he2016resnet}, Dense Convolutional Network (DenseNet) \cite{huang2019densenet}, Efficient Network (EfficientNet) \cite{tan2020efficientnet} and many others.

\subsubsection{Vision Transformers (ViTs)}
Transformer models, which were originally proposed for NLP tasks, have also found valuable applications in CV. For instance, the ViT model \cite{dosovitskiy2021vit} can capture intricate relationships and dependencies across the entire image. This is achieved by leveraging the Transformer architecture and treating images as sequences of smaller patches. Each image patch undergoes a process of flattening into a vector, followed by passage through an embedding layer. The embedding layer enriches the flattened image patches, providing a more expressive and continuous representation. Next, positional encodings are incorporated into the embeddings, conveying information about the spatial arrangement of the image patches. A distinctive feature of ViTs is the introduction of a special token designed to capture global information about the entire image. This special token has an associated learnable token embedding, represented by a vector with its unique set of parameters. ViTs have achieved notable success in semantic segmentation \cite{ranftl2021visiontf}, anomaly detection \cite{mishra2021vtadl}, medical image classification \cite{manzari2023medvit} and even outperformed CNNs in some cases \cite{tyagil2021detpneum, xin2022transfskin}. 

\section{Vision-Language Models (VLMs)}\label{sec:vlm}
Many real-world situations inherently involve a variety of data \emph{modalities}.  For example, autonomous cars must process information from various sensors like cameras, RADAR, LiDAR, and/or GPS to ensure safe and effective navigation \cite{parekh2022autocars}. Similarly, in cancer care, the fusion of radiology images with genomic data, digitized histopathology slides, and clinical reports has the potential to improve diagnosis, treatment planning, and post-treatment surveillance \cite{boehm2021advanceoncology, waqas2023multimodaloncology, Mohsan2023}. This motivated the development of VLMs,  which can handle and understand both NLP and CV data simultaneously. 

\subsection{Model Architecture} 

\subsubsection{Single- vs. Dual-Stream VLMs}Based on how different data modalities are fused together in VLMs, they are generally categorized into two groups \cite{chen2022vlp}: (1) \emph{single-stream} (e.g., VisualBERT \cite{li2019visualbert} and UNITER or UNiversal Image-TExt Representation Learning \cite{chen2019uniter}), and (2) \emph{dual-stream} models (e.g., ViLBERT Vision-and-Language BERT \cite{lu2019vilbert} and CLIP or Contrastive Language-Image Pre-training \cite{radford2021clip}). 

\paragraph{Single-Stream Models} A single-stream VLM adopts an efficient architecture for processing both visual and textual information within a unified module. This architecture incorporates an early fusion of distinct data modalities, where feature vectors from various data sources are concatenated into a single vector (e.g., MedViLL \cite{moon2022medvill}). Subsequently, this combined representation is fed into a single stream. One notable advantage of the single-stream design is its parameter efficiency, achieved by employing the same set of parameters for all modalities. This not only simplifies the model but also contributes to computational efficiency during both training and inference phases \cite{chen2022vlp}.

\paragraph{Dual-Stream Models} A dual-stream VLM extracts visual and textual representations separately in parallel streams that do not share parameters. This architecture usually has higher computational complexity than single-stream architectures. Visual features are generated from pre-trained \emph{vision encoders}, such as CNNs or ViTs, and textual features are obtained from pre-trained \emph{text encoders}, usually based on a Transformer architecture (e.g., PubMedCLIP \cite{eslami2023pubmedclip}). Both features are then fed into a \emph{multimodal fusion module}, often leveraging attention mechanisms, to integrate information from both data modalities and to learn cross-modal representations. This late fusion approach allows for more intricate interactions between visual and textual information, enabling the model to capture complex cross-modal dependencies. However, it comes at the cost of increased computational complexity compared to single-stream architecture.

\subsubsection{Encoder vs. Encoder-Decoder VLMs} The learned cross-modal representations can be optionally processed by a \emph{decoder} before producing the final output. Consequently, VLMs are classified into two groups:  (1) \emph{encoder-only} (e.g., ALIGN (A Large-scale ImaGe and Noisy-text embedding) \cite{jia2021align}), and (2) \emph{encoder-decoder} models  (e.g., SimVLM (Simple Visual Language Model) \cite{wang2022simvlm}). 

\paragraph{Encoder-only Models} These models are advantageous in scenarios where the primary objective is efficient representation learning. They often exhibit streamlined processing and reduced computational complexity, making them suitable for tasks requiring compact and informative representations. However, these models might lack the capability to generate intricate and detailed outputs, limiting their use in tasks demanding nuanced responses or creative generation.

\paragraph{Encoder-Decoder Models} These models offer the flexibility to generate complex and diverse outputs, making them well-suited for tasks like image captioning, translation, or any application requiring creative responses. The decoding step allows for the transformation of joint representations into meaningful outputs. However, this versatility comes at the cost of increased computational load and complexity. 

\subsection{Model Training}
\subsubsection{Transfer Learning}
A commonly adopted strategy in ML is to employ pre-trained models and customize them to specific downstream tasks --- a method commonly known as transfer learning. This process typically involves fine-tuning the model's parameters using smaller task-specific datasets to address the intricacies of the target task \cite{bommasani2022opportunities}. Transfer learning can also be considered as the process of starting the parameters optimization for a task using a set of already optimized parameters on another task instead of using random initialization. Transfer learning may involve some modification of the original model's architecture. This can include modifications to the final layers or the introduction of new layers, such as classification or regression layers, tailored to meet the specific requirements of the task at hand \cite{bommasani2022opportunities}. The underlying idea is to adapt the pre-trained model to the specifics of the new task while retaining the knowledge it gained during the initial pre-training. 

\subsubsection{Curriculum Learning}
Curriculum learning presents an innovative approach when dealing with tasks or data exhibiting a natural progression or hierarchy. This method involves strategically presenting training examples or tasks in a designed order, typically based on difficulty or complexity measures \cite{soviany2021currlearn}. The recent medical VLM, LLaVa-Med \cite{li2023llavamed}, adopts curriculum learning during its training phase. This allows the model to learn gradually, starting with simpler examples and progressing to more intricate ones. This orchestrated learning sequence enhances the model's adaptability and performance. 

\subsubsection{Self-Supervised Learning (SSL)} SSL is a fundamental paradigm in training VLMs, offering a powerful alternative to traditional supervised learning by allowing models to generate their own labels from the data \cite{rani2023selfsupervised}. This is particularly beneficial when obtaining large amounts of labeled data is challenging or expensive. In self-supervised learning for VLMs, the models formulate tasks that leverage inherent structures within the data, enabling them to learn meaningful representations across modalities without explicit external labels. Contrastive learning, masked language modeling, and masked image modeling (described in the following sub-section) are examples of self-supervised learning tasks.

\subsubsection{Pre-Training Process and Tasks} 
The pre-training process plays a pivotal role in equipping VLMs with a foundational understanding of the intricate interplay between visual and textual data. A prevalent strategy involves intensive pre-training on datasets where images/videos are paired with their corresponding textual descriptions. During pre-training, various tasks guide the model in learning versatile representations for downstream tasks.

\paragraph{Contrastive Learning (CL)} CL encourages the model to learn meaningful representations by contrasting positive pairs with negative pairs of both visual and textual data \cite{li2021albef}. During CL, the model is trained to map both positive and negative pairs into a shared embedding space. Positive pairs consist of examples where the visual and textual content are related, such as an image paired with its corresponding textual description. Conversely, negative pairs consist of examples where the visual and textual content are unrelated, like an image paired with a randomly selected different textual description. The objective is to bring positive pairs closer together while pushing negative pairs farther apart in the shared embedding space.  Various contrastive loss functions are employed to achieve this objective, with the \emph{InfoNCE (Noise-Contrastive Estimation) loss} \cite{oord2019infonceloss} being a common choice. InfoNCE formulates a probabilistic task where the model is trained to maximize the likelihood of observing positive pairs and minimize the likelihood of observing negative pairs. The negative log-likelihood of the positive pair is used as the loss. CLIP \cite{radford2021clip} employs InfoNCE loss with cosine similarity. On the other hand, ALIGN \cite{jia2021align} uses a \emph{normalized softmax loss}. This loss computes the softmax over cosine similarities between the normalized embeddings of positive and negative pairs, aiming to boost positive similarity while diminishing negative similarities.

\paragraph{Masked Language Modeling (MLM)} MLM is a widely used task in NLP \cite{taylor1953mlm}. It was first introduced and applied in the BERT model \cite{devlin2019bert}. MLM involves randomly selecting a percentage of tokens within textual data and replacing them with a special token, often denoted as MASK. The model predicts these masked tokens by taking into account the context on both sides of them, allowing the model to grasp nuanced contextual information.  VLMs such as UNITER \cite{chen2019uniter} and VisualBERT \cite{li2019visualbert} leverage MLM for pre-training.

\paragraph{Masked Image Modeling (MIM)} Extending the idea of MLM to images gave rise to MIM \cite{xie2022simmim}. In MIM, certain patches are masked, prompting the model to predict the contents of masked regions. This process enables the model to draw context from the entirety of the image, encouraging the integration of both local and global visual features. VLMs like UNITER \cite{chen2019uniter} and ViLBERT \cite{lu2019vilbert} leverage MIM for enhanced performance.  The \emph{cross-entropy loss} is employed in MLM and MIM tasks to measure the difference between predicted and actual probability distributions for the masked elements. Additionally, MLM can be combined with MIM, allowing the reconstruction of the masked signal in one modality with support from another modality \cite{kwon2023maskedvlm}. 

\paragraph{Image-Text Matching (ITM)} ITM is another common vision-language pre-training task. Throughout the training, the model learns to map images and corresponding textual descriptions into a shared semantic space, where closely aligned vectors represent similar content in both modalities. In single-stream VLMs, the special token [CLS] represents the joint representation for both modalities. In contrast, in dual-stream VLMs, the visual and textual representations of [CLS]\textsubscript{V} and [CLS]\textsubscript{T} are concatenated. This joint representation is fed into a fully-connected layer followed by the sigmoid function, predicting a score indicating match or mismatch \cite{chen2022vlp}. Models like CLIP \cite{radford2021clip}, ALBEF (ALign the image and text representations BEfore Fusing) \cite{li2021albef}, and METER \cite{dou2022meter} leverage ITM during pre-training.

\paragraph{Combining Multiple Tasks}
In VLM pre-training, multiple tasks are often combined in a unified framework, allowing models to grasp nuanced contextual information across modalities. The final loss function can combine contrastive loss, cross-entropy loss for masked token prediction, and other task-specific losses. This comprehensive pre-training approach equips VLMs with versatile representations for diverse downstream tasks. For example, ALBEF \cite{li2021albef} adopts a comprehensive pre-training objective encompassing three tasks: CL, MLM, and ITM. The overall loss is then computed as the sum of these individual components.

\subsubsection{Fine-Tuning Techniques}
Following the training, a common practice involves \emph{fine-tuning} VLMs on smaller datasets tailored to specific downstream tasks.

\paragraph{Supervised Fine-Tuning (SFT)} Before employing SFT, the VLM is pre-training on an extensive image-text dataset, establishing a foundational understanding of the complex relationship between visual and textual representations. SFT involves meticulous fine-tuning on a more focused dataset, curated to match the nuances of the targeted application. This dual-phase strategy, encompassing broad pre-training and task-specific fine-tuning, enables the model to benefit from large-scale generalization while seamlessly adapting to the intricacies of particular applications \cite{ouyang2022training}.

\paragraph{Reinforcement Learning from Human Feedback (RLHF)} RLHF is a distinct fine-tuning approach employed to enhance VLMs through the incorporation of human preferences during fine-tuning \cite{ouyang2022training, lambert2022illustrating, ziegler2020finetuning}. RLHF initiates with an initial model, incorporating human-generated rankings of its outputs to construct a detailed reward model. In contrast to traditional reinforcement learning (RL) \cite{sutton1998rlintro, coronato2020rl}, which relies solely on environmental interactions, RLHF strategically integrates human feedback. This human-in-the-loop approach provides a more nuanced and expert-informed methodology, allowing for the fine-tuning of VLMs in alignment with human preferences, ultimately leading to improved model outcomes.

\paragraph{Instruction Fine-Tuning (IFT)} IFT refers to the process of refining a pre-trained language model by providing specific instructions or guidance tailored to a particular task or application \cite{ren2024learning-IFT}. This process typically involves exposing the model to examples or prompts related to the desired instructions and updating its parameters based on the feedback received during this task-specific training phase. Medical VLM, RaDialog \cite{pellegrini2023radialog}, employs this fine-tuning technique.

\subsection{Parameter-Efficient Fine-Tuning (PEFT)}
In this section, we explore strategies for adapting VLMs while keeping the model's parameters frozen and only updating newly added layers.
In recent years, PEFT has gained prominence, encompassing various techniques and strategies that aim to make the most effective use of parameters during the fine-tuning process, particularly in scenarios with limited labeled data for the target task.  The main strategy of PEFT is incorporating task-specific parameters, known as \emph{adapters}, into a pre-trained model while retaining its original parameters. The architecture of adapter modules typically incorporates a bottleneck structure, projecting original features into a reduced dimension, applying non-linearity, and then projecting back to the original dimension. This thoughtful design ensures parameter efficiency by limiting the number of added parameters per task. Integrated after each layer of the pre-trained model, adapter modules capture task-specific details while preserving shared parameters, enabling the model's seamless extension to new tasks without significant interference with previously acquired knowledge. 

\subsubsection{Low-Rank Adaptation (LoRA)} LoRA is a common adapter-based method is \cite{hu2022lora}. The adaptation process involves fine-tuning two smaller low-rank matrices that are decompositions of the larger weight matrix of the pre-trained model. These smaller matrices constitute the LoRA adapter modules, and the approach focuses on making low-rank modifications to adapt the model for specific tasks efficiently. Pre-trained LLMs that are part of medical VLMs architecture are often fine-tuned using LoRA (e.g., Visual Med-Alpaca \cite{shu2023vismedalpaca} and RaDialog \cite{pellegrini2023radialog}).

\subsubsection{Prompt Tuning} Prompt tuning involves creating continuous vector representations as input hints \cite{lester2021prompttuning}. This allows the model to generate effective prompts during training dynamically. Continuous refinement of prompts significantly enhances the model's ability to generate accurate and contextually relevant responses. This iterative process allows the model to adapt its behavior based on an evolving understanding of the task. VLMs like Qwen-VL and InstructBLIP used prompt tuning \cite{bai2023qwenvl, dai2023instructblip}.

\subsubsection{Prefix Token Tuning} Prefix token tuning adds task-specific vectors to the input, specifically to the initial tokens known as \emph{prefix tokens}, to guide the model's behavior for a given task \cite{li2021prefixtuning}. For instance, VL-T5 utilized different prefixes for questions from various datasets \cite{cho2021vlt5} . These vectors can be trained and updated independently while keeping the remaining pre-trained model parameters frozen. Prefix token tuning allows for task-specific adaptation without compromising the pre-trained knowledge encoded in the majority of the model's parameters.

\subsection{In-Context Learning}
In this section, we explore strategies for adapting VLMs using the context only, keeping the model's parameters (and PEFT/LoRA adapters, if any) frozen. In our settings, in-context learning may be considered as using LLMs or VLMs for inference only. 

\subsubsection{Prompt Engineering}  Prompt engineering is a technique that involves enhancing a large pre-trained model with task-specific instructions, referred to as \emph{prompts}, to tailor the model's output for specific tasks \cite{gu2023promptvlm}. Examples include instructing the model to generate a radiology report for a specific image (e.g., RAMM \cite{pellegrini2023radialog}). Prompt engineering can also expose the VLM to a sequence of interconnected examples or prompts, guiding it to a desired output. Another approach incorporates progressively structured instructions or questions, refining focus and enhancing the model's ability to generate coherent and contextually relevant responses \cite{gu2023promptvlm}.

\subsubsection{Retrieval augmented generation (RAG)} RAG is a form of prompt engineering that involves strategically crafting prompts for both retrieval and generation phases, allowing for an adaptive and efficient process that leverages external knowledge sources to enhance generative tasks. While the original concept of RAG was developed in the context of NLP \cite{lewis2020rag}, the principles behind retrieval and generation can be extended to multimodal learning \cite{zhao2023multimodalrag}, including VLMs. RAG has been used in medical VLMs for tasks like VQA (e.g., RAMM \cite{yuan2023ramm}) and RG (e.g., CXR-RePaiR-Gen \cite{ranjit2023cxrrepairgen}). RAG begins with a retrieval component, which is usually a pre-trained model designed for information retrieval. This versatile component excels in extracting pertinent information from extensive datasets, catering to various modalities such as images, text, codes, video, or audio when presented with diverse inputs \cite{zhao2023multimodalrag}. Following the retrieval phase, the model returns a set of contexts related to the given input. The second component is a generative LLM. This component takes the input and the retrieved context and generates the final output. The generated output is conditioned not only on the input but also on the information extracted from the retrieved context. An intrinsic advantage of RAG lies in its capacity to reduce the reliance on extensive labeled datasets. While the base model is typically frozen during RAG, there are instances, as seen in RAMM \cite{yuan2023ramm}, where model parameters are updated in the process.

\subsection{Downstream Tasks}
Multimodal downstream tasks leverage the acquired knowledge from pre-training VLMs to excel in diverse applications that require a joint understanding of visual and textual data. 

\subsubsection{Report Generation (RG)} RG is a prominent example of a typical medical VLM task, which centers on creating a comprehensive summary report of visual data. RG plays a crucial role in automatically summarizing diagnostic imaging results and reducing the workload of report writing \cite{monshi2020radreportsurvey, ting2023mrgsurvey, Mohsan2023}. For instance, in radiology, a report generation system could analyze a set of medical images such as X-rays, CT scans, or MRIs and generate a detailed report summarizing the observed abnormalities, their locations, and potential implications for diagnosis or treatment \cite{liu2023rrgreview}. A radiology report usually has several sections: (1) \emph{Examination} (type of exam), (2) \emph{Indication} (reasons for the examination), (3) \emph{Comparison} (prior exams), (4) \emph{Technique} (scanning method) (5) \emph{Findings} (detailed observations made by a radiologist), and (6) \emph{Impression} (summary of the major findings) \cite{mabotuwana2020radreportfram}. In the context of RG, VLMs are usually designed to generate \emph{Findings} and \emph{Impression} sections \cite{thawkar2023xraygpt}. Currently, VLMs tailored for RG are predominantly utilized for radiology images, with lesser application in other medical imaging domains such as pathology \cite{sengupta2023pathologyrg}, robotic surgery \cite{Xu2021rginsurgery}, and ophthalmology \cite{li2022rginophthalmology}.

\subsubsection{Visual Question Answering (VQA)} 
VQA is another important visual-language understanding task, where the model needs to comprehend images or videos and the posed question to provide a relevant and accurate response  \cite{antol2015vqa}. 
The spectrum of questions encountered in VQA is broad, encompassing inquiries about the presence of specific objects, their locations, or distinctive properties within the image. In the medical context \cite{lin2023medvqa}, this may involve questions regarding the presence of medical conditions or abnormalities, such as ``What abnormality is seen in the image?'' \cite{ionescu2021vqamed21} or ``Is there gastric fullness?'' \cite{lau2018vqarad}. Other queries may delve into details like the imaging method used \cite{abacha2019vqamed19}, the organ system involved \cite{lau2018vqarad}, or the presence of specific anatomical structures \cite{liu2021slake}. 

Questions in VQA fall into two categories. \emph{Open-ended questions} elicit responses in the form of phrases or sentences, fostering detailed and nuanced answers \cite{thawkar2023xraygpt}. On the other hand, \emph{closed-ended questions} are designed to prompt limited responses, often with predetermined options, such as a short list of multiple choices, a yes/no response, or a numeric rating \cite{bazi2023vqamedim}. The task of VQA is commonly approached as either a classification task, a generation task, or both \cite{lin2023medvqa}. In the classification approach, models select the correct answer from a predefined set, while in the generation task, models produce free-form textual responses unconstrained by predefined options.

\subsubsection{Other Tasks}
Beyond VQA and RG, a spectrum of VLM tasks exist for the vision-language understanding \cite{chen2022vlp}. For instance, \emph{referring expression comprehension} entails a model locating the specific area or object in an image that the given phrase or sentence refers to \cite{zhang2018gre}. \emph{Visual commonsense reasoning} involves answering questions about an image, typically presented in a multiple-choice format, and justifying the answer based on the model's understanding of the image and common sense knowledge \cite{zellers2019vcr}. \emph{Vision-language retrieval} focuses on either generating or retrieving relevant information from images using textual data, or vice versa, obtaining information from text using visual data \cite{zhen2019dscretrival}. In the context of \emph{visual captioning}, the model's role is to generate a concise, text-based description of either an image \cite{sharma2023captioning}.
It is worth highlighting that some of these tasks can seamlessly transition from images to videos, showcasing the adaptability and versatility of VLMs across diverse visual contexts \cite{gan2022vlmsurvey}.

\begin{table}
\caption{A list of datasets used for developing medical VLMs.}\label{tab:datasets}
    \centering
    \scalebox{0.62}{\begin{tabular}{ccccc}
    \hline \hline
    \textbf{Dataset}& \textbf{\# image-text pairs} & \textbf{\# QA pairs}  & \textbf{Other components} & \textbf{Link} \\
    \hline 
    \textbf{ROCO}  & \multirow{2}{*}{$81,825$} & \multirow{2}{*}{--}  & \multirow{2}{*}{--} &  \multirow{2}{*}{\href{https://github.com/razorx89/roco-dataset}{GH}}\\
    \cite{pelka2018roco} &  &  &  &  \\
     &  &  &  &  \\
    \textbf{MIMIC-CXR} & \multirow{2}{*}{$377,110$} & \multirow{2}{*}{--}  & \multirow{2}{*}{--} &  \multirow{2}{*}{\href{https://www.physionet.org/content/mimic-cxr/2.0.0/}{PN}}\\
    \cite{johnson2019mimiccxr} &  &  &  &  \\
     &  &  &  &  \\
    \textbf{MIMIC-CXR-JPG} & \multirow{2}{*}{$377,110$} & \multirow{2}{*}{--}  & \multirow{2}{*}{pathology labels} &  \multirow{2}{*}{\href{https://physionet.org/content/mimic-cxr-jpg/2.0.0/}{PN}}\\
    \cite{johnson2019mimiccxrjpg} &  &   &  &  \\
     &  &  &  &  \\
    \textbf{MIMIC-NLE} & \multirow{2}{*}{$38,003$} & \multirow{2}{*}{--}  & diagnosis labels, &  \multirow{2}{*}{\href{https://github.com/maximek3/MIMIC-NLE}{GH}}\\
    \cite{kayser2022mimicnle} &  &   &  evidence labels &  \\
     &  &  &  &  \\
    \textbf{CXR-PRO} & \multirow{2}{*}{--} & \multirow{2}{*}{--}  & $374,139$ radiographs and &  \multirow{2}{*}{\href{https://physionet.org/content/cxr-pro/1.0.0/}{PN}}\\
    \cite{ramesh2022cxrredone} &  &   & $374,139$ reports but not paired &  \\
      &  &  &  &  \\
    \textbf{MS-CXR} & \multirow{2}{*}{$1,162$} & \multirow{2}{*}{--} & \multirow{2}{*}{bounding box annotations} &  \multirow{2}{*}{\href{https://physionet.org/content/ms-cxr/0.1/}{PN}}\\
    \cite{boecking2022mscxr} &  &   &  &  \\
     &  &  &  &  \\
    \textbf{IU-Xray or Open-I}  & \multirow{2}{*}{$7,470$} & \multirow{2}{*}{--}  & \multirow{2}{*}{labels} &  \multirow{2}{*}{\href{https://openi.nlm.nih.gov/}{Site}}\\
    \cite{demnerfushman2015openi} &  &  &  &  \\
      &  &  &  &  \\
    \textbf{MedICaT}  & \multirow{2}{*}{$224,567$} & \multirow{2}{*}{--}  & annotations;  inline &  \multirow{2}{*}{\href{https://github.com/allenai/medicat}{GH}}\\
    \cite{subramanian2020medicat} &  &  & references to ROCO figures &  \\
      &  &  &  &  \\
    \textbf{PMC-OA}  & \multirow{2}{*}{$1,650,000$}  & \multirow{2}{*}{--}  & \multirow{2}{*}{--} &  \multirow{2}{*}{\href{https://huggingface.co/datasets/axiong/pmc_oa}{HF}} \\
    \cite{lin2023pmcclip} &  &  &  &  \\
      &  &  &  &  \\
    \textbf{SLAKE} & \multirow{2}{*}{--} & \multirow{2}{*}{$14,028$}  & $642$ annotated images,  &  \multirow{2}{*}{\href{https://www.med-vqa.com/slake/}{Site}}\\
     \cite{liu2021slake} &  &   &  $5,232$ medical triplets &  \\
       &  &  &  &  \\
    \textbf{VQA-RAD}  & \multirow{2}{*}{--}  &  \multirow{2}{*}{$3,515$} &  \multirow{2}{*}{$315$ radiology images}  &  \multirow{2}{*}{\href{https://osf.io/89kps/}{Site}} \\
    \cite{lau2018vqarad} &  &  &  &  \\
      &  &  &  &  \\
    \textbf{PathVQA}  & \multirow{2}{*}{--}  & \multirow{2}{*}{$32,799$}  & \multirow{2}{*}{$4,998$ pathology images} &  \multirow{2}{*}{\href{https://github.com/UCSD-AI4H/PathVQA}{GH}} \\
    \cite{he2020pathvqa} &  &  &  &  \\
      &  &  &  &  \\
    \textbf{VQA-Med 2019} & \multirow{2}{*}{--}  & \multirow{2}{*}{$15,292$}  & \multirow{2}{*}{$4,200$ radiology images} &  \multirow{2}{*}{\href{https://github.com/abachaa/VQA-Med-2019}{GH}} \\
    \cite{abacha2019vqamed19}  &  &  &   &  \\
      &  &  &  &  \\
    \textbf{VQA-Med 2020} & \multirow{2}{*}{--}  & \multirow{2}{*}{$5,000$}  & $5,000$ radiology images for VQA;  &  \multirow{2}{*}{\href{https://github.com/abachaa/VQA-Med-2020}{GH}} \\
    \cite{abacha2020vqamed20}  &  &  & images and questions for VQG  &  \\
      &  &  &  &  \\
    \textbf{VQA-Med 2021} & \multirow{2}{*}{--}  & \multirow{2}{*}{$5,500$}  & $5,500$ radiology images for VQA;  &  \multirow{2}{*}{\href{https://github.com/abachaa/VQA-Med-2021}{GH}} \\
    \cite{ionescu2021vqamed21}  &  &  & images and questions for VQG  &  \\
      &  &  &  &  \\
    \textbf{EndoVis 2017} & \multirow{2}{*}{--}  & \multirow{2}{*}{$472$}  & bounding box annotations;  &  \multirow{2}{*}{\href{https://github.com/longbai1006/Surgical-VQLA}{GH}} \\
    \cite{allan2019endovis17}  &  &  & 97 frames  &  \\
      &  &  &  &  \\
    \textbf{EndoVis 2018} & \multirow{2}{*}{--}  & \multirow{2}{*}{$11,783$}  & bounding box annotations;  &  \multirow{2}{*}{\href{https://github.com/longbai1006/Surgical-VQLA}{GH} + \href{https://endovissub2018-roboticscenesegmentation.grand-challenge.org/Data/}{Site}}\\
    \cite{allan2020endovis18}  &  &  & 2,007 frames  &  \\
     \\ \hline
      \multicolumn{5}{l}{Note: Abbreviations used are: GH - GitHub, HF - Hugging Face, and PN - PhysioNet} \\
    \hline \hline
    \end{tabular}}
\end{table}

\section{Medical VLMs}\label{sec:medical_vlms}

\subsection{Medical Datasets for VLMs}\label{subsec:datasets}

The adaptation of VLMs to various medical tasks is achieved through their pre-training and fine-tuning using specialized task-specific datasets. Below is the list of vision-language datasets available in the public domain that contain medical image-text pairs or question-answer (QA) pairs. Most of them are employed by medical VLMs described in \cref{subsec:models} for pre-training, fine-tuning, and evaluating VQA and RG tasks. The comparative analysis of these datasets is presented in  \cref{tab:datasets}. The last column in \cref{tab:datasets} provides a link to the source of the data on the web with the following abbreviations: GH - GitHub, PN - PhysioNet, and HF - Hugging Face.

\subsubsection{Radiology Objects in Context (ROCO)} ROCO is a dataset composed of image-caption pairs extracted from the open-access biomedical literature database PubMed Central (PMC) \cite{pelka2018roco}. ROCO is stratified into two categories: radiology and out-of-class.  The radiology group includes $81,825$ radiology images, including computer tomography (CT), ultrasound, x-ray, fluoroscopy, positron emission tomography (PET), mammography, magnetic resonance imaging (MRI), angiography, and PET-CT. The out-of-class group has $6,127$ images, including synthetic radiology images, clinical photos, portraits, compound radiology images, and digital art. Each image is accompanied by a corresponding caption, keyword,  Unified Medical Language System (UMLS) semantic types (SemTypes), UMLS concept unique identifiers (CUIs), and a download link. To facilitate model training, the dataset is randomly split into a training set ($65,460$ radiology and $4,902$ out-of-class images), a validation set ($8,183$ radiology and $612$ out-of-class images), and a test set ($8,182$ radiology and $613$ out-of-class images) using an 80/10/10 split ratio, respectively.

\subsubsection{Medical Information Mart for Intensive Care - Chest X-Ray (MIMIC-CXR)} 
MIMIC-CXR collection encompasses $377,110$ chest X-rays paired with $227,835$ associated free-text radiology reports \cite{johnson2019mimiccxr}. The dataset is derived from de-identified radiographic studies conducted at the Beth Israel Deaconess Medical Center in Boston, MA. Each imaging study within the MIMIC-CXR dataset consists of one or more images, typically featuring lateral and from back-to-front (posteroanterior, PA) views in Digital Imaging and Communications in Medicine (DICOM) format. 
    
\subsubsection{MIMIC-CXR-JPG}
MIMIC-CXR-JPG \cite{johnson2019mimiccxrjpg} is a pre-processed variant of the MIMIC-CXR dataset \cite{johnson2019mimiccxr}. In this version, the original $377,110$ images are converted into compressed JPG format. The $227,827$ reports associated with these images are enriched with labels for various common pathologies. The labels are derived from the analysis of the impression, findings, or final sections of the radiology reports, facilitated by the use of NegBio \cite{peng2017negbio} and CheXpert (Chest eXpert) \cite{irvin2019chexpert} tools.
    
\subsubsection{MIMIC-NLE} 
MIMIC-NLE dataset is specifically designed for the task of generating natural language explanations (NLEs) to justify predictions made on medical images, particularly in the context of thoracic pathologies and chest X-ray findings \cite{kayser2022mimicnle}. The dataset consists of $38,003$ image-NLE pairs or $44,935$ image-diagnosis-NLE triplets, acknowledging instances where a single NLE may explain multiple diagnoses. NLEs are extracted from MIMIC-CXR \cite{johnson2019mimiccxr} radiology reports. The dataset exclusively considers X-ray views from front-to-back (anteroposterior, AP) and back-to-front (posteroanterior, PA). All NLEs come with diagnosis and evidence (for a diagnosis) labels. The dataset is split into the training set with $37,016$ images, a test set with $273$ images, and a validation set with $714$ images.

\subsubsection{CXR with Prior References Omitted (CXR-PRO)}
CXR-PRO dataset is derived from MIMIC-CXR \cite{johnson2019mimiccxr}. The dataset consists of $374,139$ free-text radiology reports containing only the impression sections \cite{ramesh2022cxrredone}. It also incorporates associated chest radiographs; however, the radiology reports and chest X-rays are not paired. This dataset is designed to mitigate the problem of hallucinated references to prior reports often generated by radiology report generation ML models. The omission of prior references in this dataset aims to provide a cleaner and more reliable dataset for radiology RG.

\subsubsection{Indiana University chest X-rays (IU-Xray)}
IU-Xray dataset, also known as the \emph{Open-I} dataset, is accessible through the National Library of Medicine's Open-i service \cite{demnerfushman2015openi}. The dataset originates from two hospital systems within the Indiana Network for Patient Care database. This dataset comprises $7,470$ DICOM chest X-rays paired with $3,955$ associated radiology reports. The reports typically include sections such as indications, findings, and impressions, and they are manually annotated using MeSH and RadLex (Radiology Lexicon) codes to represent clinical findings and diagnoses. Throughout this review, we will refer to the dataset interchangeably as \emph{IU-Xray} and \emph{Open-I}, maintaining consistency with the nomenclature used in related literature.

\subsubsection{Medical Images, Captions, and Textual References (MedICaT)}
MedICaT dataset contains $217,060$ figures from $131,410$ open-access PMC papers focused on radiology images and other medical imagery types \cite{subramanian2020medicat}. Excluding figures from ROCO \cite{pelka2018roco}, the dataset integrates inline references from the S2ORC (Semantic Scholar Open Research Corpus) \cite{lo2020s2orc} corpus, establishing connections between references and corresponding figures. Additionally, the inline references to ROCO figures are provided separately. MedICaT also contains $7,507$ subcaption-subfigure pairs with annotations derived from $2,069$ compound figures. 

\subsubsection{PubMedCentral’s OpenAccess (PMC-OA)}
PMC-OA dataset comprises $1.65$ M image-caption pairs, derived from PMC papers \cite{lin2023pmcclip}. It encompasses a variety of diagnostic procedures, including common ones such as ultrasound, MRI, PET, and radioisotope, and rarer procedures like mitotic and fMRI. Additionally, the dataset covers a broad spectrum of diseases, with induced cataracts, ear diseases, and low vision being among the most frequently represented conditions.

\subsubsection{MS-CXR}
MS-CXR dataset contains image bounding box labels paired with radiology findings, annotated and verified by two board-certified radiologists \cite{boecking2022mscxr}. The dataset consists of $1,162$ image-text pairs of bounding boxes and corresponding text descriptions.  The annotations cover 8 different cardiopulmonary radiological findings and are extracted from MIMIC-CXR \cite{johnson2019mimiccxr} and REFLACX (Reports and Eye-tracking data For Localization of Abnormalities in Chest X-rays) \cite{lanfredi2022reflacx} (based on MIMIC-CXR) datasets. The findings include atelectasis, cardiomegaly, consolidation, edema, lung opacity, pleural effusion, pneumonia, and pneumothorax. 

\subsubsection{Semantically-Labeled Knowledge-Enhanced (SLAKE)}
SLAKE is an English-Chinese bilingual dataset \cite{liu2021slake}. It contains $642$ images, including $12$ diseases and $39$ organs of the whole body. Each image is meticulously annotated with two types of visual information: masks for semantic segmentation and bounding boxes for object detection. The dataset includes a total of $14,028$ QA pairs, categorized into vision-only or knowledge-based types and labeled accordingly, encompassing both open- and closed-ended questions.  Moreover, SLAKE incorporates $5,232$ medical knowledge triplets in the form of $<head, relation, tail>$, where $head$ and $tail$ denote entities (e.g., organ, disease), and $relation$ signify the relationship between these entities (e.g., function, treatment). An illustrative example of such a triplet is $<$pneumonia, location, lung$>$.

\subsubsection{VQA-RAD}
VQA-RAD dataset contains $104$ head axial single-slice CTs or MRIs, $107$ chest x-rays, and $104$ abdominal axial CTs \cite{lau2018vqarad}.  The images are meticulously chosen from MedPix, an open-access online medical image database, ensuring each image corresponds to a unique patient. Furthermore, every selected image has an associated caption and is deliberately devoid of any radiology markings. Every caption provides details about the imaging plane, modality, and findings generated and reviewed by expert radiologists. Also, VQA-RAD contains $3,515$ QA pairs, with an average of 10 questions per image. Among them, $1,515$ are free-form questions and answers, allowing for unrestricted inquiry. Additionally, $733$ pairs involve rephrased questions and answers, introducing linguistic diversity. Another $1,267$ pairs are framed, featuring questions presented in a structured format, offering consistency and systematic evaluation. Additionally, QA pairs are split into $637$ open-ended and $878$ closed-ended types. Within the closed-ended group, a predominant focus is on yes/no questions. 

\subsubsection{PathVQA}
PathVQA is a dataset that encompasses $4,998$ pathology images accompanied by a total of $32,799$ QA pairs derived from these images \cite{he2020pathvqa}. The images are sourced from pathology books: ``Textbook of Pathology'' and ``Basic Pathology'', and the digital library ``Pathology Education Informational Resource''. Out of all QA pairs, $16,465$ are of the open-ended type, while the remaining pairs are of the closed-ended yes/no type. On average, each image is associated with $6.6$ questions, which cover a broad spectrum of visual contents, encompassing aspects such as color, location, appearance, shape, etc. 

\subsubsection{VQA-Med 2019}
VQA-Med 2019 dataset contains $4,200$ radiology images obtained from MedPix, an open-access online medical image database, and $15,292$ QA pairs \cite{abacha2019vqamed19}. The training set consists of $3,200$ images and $12,792$ QA pairs, with each image having 3 to 4 associated questions. The validation set includes $500$ images and $2,000$ QA pairs, and the test set comprises $500$ images and $500$ QA pairs. The questions are mainly about modality, imaging plane, organ system, and abnormality.

\subsubsection{VQA-Med 2020}
VQA-Med 2020 dataset contains $5,000$ radiology images obtained from MedPix, an open-access online medical image database, and $5,000$ QA pairs \cite{abacha2020vqamed20}. The training set consists of $4,000$ images and $4,000$ QA pairs. The validation set comprises $500$ images and $500$ QA pairs, and the test set includes $500$ images and $500$ QA pairs. The questions are focused on abnormalities present in the images. Additionally, the dataset contains radiology images and questions for the Visual Question Generation (VQG) task. The training set consists of $780$ images and $2,156$ associated questions. The validation set comprises $141$ images with $164$ questions, and the test set includes $80$ images.

\subsubsection{VQA-Med 2021}
VQA-Med 2021 dataset contains $5,500$ radiology images obtained from MedPix, an open-access online medical image database, and $5,500$ QA pairs \cite{ionescu2021vqamed21}. The training set consists of $4,500$ images and $4,5000$ QA pairs. The validation set comprises $500$ images and $500$ QA pairs, and the test set includes $500$ images and $500$ QA pairs. The questions are focused on abnormalities present in the images. Similarly to VQA-Med 2019, the dataset also contains radiology images and questions for the VQG task. The validation set comprises $85$ images with $200$ questions, and the test set includes $100$ images.

\subsubsection{Endoscopic Vision (EndoVis) 2017}
EndoVis 2017 dataset contains 5 robotic surgery videos (two videos with 8 frames each, one with 18, one with 14, and one with 39 frames) from the MICCAI (Medical Image Computing and Computer Assisted Interventions) Endoscopic Vision 2017 Challenge \cite{allan2019endovis17}. It also includes 472 QA pairs with bounding box annotations. These QA pairs are carefully crafted to involve specific inquiries related to the surgical procedure. Examples of questions include queries such as "What is the state of prograsp forceps?" and ``Where is the large needle driver located?''. The inclusion of bounding box annotations enhances the dataset's utility for tasks such as object detection or answer localization. 

\subsubsection{EndoVis 2018}
EndoVis 2018 dataset contains 14 robotic surgery videos ($2,007$ frames in total) from the MICCAI Endoscopic Vision 2018 Challenge \cite{allan2020endovis18}. It also includes $11,783$ QA pairs regarding organs, surgical tools, and organ-tool interactions. When the question is about organ-tool interactions, the bounding box will contain both the organ and the tool.

\subsection{VLM Evaluation Metrics}\label{subsec:metrics}
This section delves into the evaluation process of medical VLMs. The initiation of this process involves meticulously selecting benchmark datasets and defining evaluation metrics tailored to the specific vision-language tasks at hand. 

\subsubsection{Evaluation Metrics for Report Generation}
The prevalent benchmark datasets for medical RG are MIMIC-CXR \cite{johnson2019mimiccxr} and Open-I \cite{demnerfushman2015openi}. For more information on these datasets, see \cref{subsec:datasets}. Several metrics are used to evaluate the effectiveness of VLMs on RG tasks. The more frequently used metrics are outlined below.

\paragraph{Bilingual Evaluation Understudy (BLEU)} The BLEU score was originally designed for machine translation evaluation, but it has been adapted for RG and even VQA in a modified form. BLEU provides a quantitative measure of how well the machine-generated text aligns with human-generated reference text \cite{papineni2002bleu}. First, the precision of different \emph{n-grams}, which are consecutive sequences of $n$ words, is calculated using the formula:
    \begin{align}
    \text{Precision}(n) = \frac{\# \text{overlapping n-grams}}{\# \text{all n-grams in a model-generated text}}, 
    \end{align}
    
    where `overlapping n-grams' refer to n-grams in the model-generated text that share common elements with at least one n-gram in the reference text. To ensure the precision score remains robust and is not disproportionately affected by repeated n-grams in the model-generated text, a modification known as clipping is often introduced. This process involves capping the count of each n-gram in the model-generated text to a maximum count. This maximum count is determined by the highest count observed in any single reference text for the same n-gram.  The final BLEU-n score is defined as:
    \begin{align} \label{eq-BLEU-n}
    \text{BLEU-n} = BP \times \frac{1}{n}\exp\left(\sum_{k=1}^n \log \left[\text{Precision(k)}\right]\right).    
    \end{align}
    In eq. \ref{eq-BLEU-n}, $BP$ is referred to as the brevity penalty and is calculated as: 
    \begin{align}
        BP = \begin{cases} 
    1 & \text{if } c \geq r \\
    e^{(1 - r/c)} & \text{if } c < r,
   \end{cases}
    \end{align}
   where $c$ is the length of the model-generated text, and $r$ is the length of the reference text. It is common to use $n=4$. The BLEU score ranges from 0 to 1, where a higher score suggests better agreement with the reference text. The overall BLEU score of the model is the average of BLEU scores for each pair of reports.

\paragraph{Recall-Oriented Understudy for Gisting Evaluation (ROUGE)} ROUGE is a set of metrics that evaluate the overlap between the model-generated text and human-generated reference text \cite{lin2004rouge}. ROUGE-n assesses the overlap of n-grams between model-generated text and reference text, and it is defined as:
   \begin{align}
       \text{ROUGE-n} = \frac{\# \text{overlapping n-grams}}{\# \text{all n-grams in a reference text}}.
   \end{align}
   
   ROUGE-L focuses on measuring the longest common subsequence between model-generated text $Y$ and reference text $X$, and it is calculated using the following relationship:
   \begin{align}
       \text{ROUGE-L} = \frac{(1+\beta^2) \times R \times P}{(R+P\times \beta^2)},
   \end{align}
   where $R={LCS(X, Y)}/{m}$, $P={LCS(X, Y)}/{n}$, $m$ is the length of $X$, $n$ is the length of $Y$, $LCS(X, Y)$ is the length of a longest common subsequence of $X$ and $Y$, and $\beta$ is a parameter that depends on the specific task and the relative importance of precision (P) and recall (R). There are other ROUGE score variants.
   The ROUGE scores range from 0 to 1, where higher scores indicate similarity between the model-generated text and the reference text.  For each ROUGE variant, the overall score of the model is the average of scores for each instance.

\paragraph{Metric for Evaluation of Translation with Explicit ORrdering (METEOR)} METEOR is an evaluation metric designed to be more forgiving than some other metrics and takes into account the fluency and meaning of the generated text  \cite{banarjee2005meteor}. The METEOR score is computed as follows:

    \begin{align}
        \text{METEOR} = \frac{10 \times P \times R}{R + 9 \times P} ( 1 - \text{Penalty})   
    \end{align}
   where 
   \begin{align}
       R &= \frac{\# \text{overlapping 1-grams}}{ \# \text{1-grams in a reference text}}, \\
       P &= \frac{\# \text{overlapping 1-grams}}{ \# \text{1-grams in a model-generated text}}, \\
       \text{Penalty} &= \frac{1}{2} \times \left(\frac{\# \text{chunks}}{ \# \text{overlapping 1-grams}}\right)^3,
   \end{align}
   and $chunks$ are groups of adjacent 1-grams in the model-generated text that overlap with adjacent 1-grams in the reference text. The METEOR score ranges from 0 to 1, with higher scores indicating better alignment between the model-generated text and the reference text. The overall METEOR score of a model is the average of scores for each instance.
   
\paragraph{Perplexity} Perplexity measures the average uncertainty of a model in predicting each word in a text \cite{hao2020preplexity}. The formula for perplexity is defined as:
    \begin{align}
        \text{Perplexity} = \exp \left( {-\frac{1}{n}\sum_{k=1}^n \ln P(w_k | w_1, w_2, \dots, w_{k-1})} \right),
    \end{align}
   where $n$ is the total number of words in the text. The value of the perplexity metric can range from  1 to $+ \infty$, and lower values signify a more accurate and confident model in capturing the language patterns within the given text.
      
\paragraph{BERTScore} BERTScore metric was initially designed for evaluating models that use BERT \cite{devlin2019bert} embeddings \cite{zhang2020bertscore}. However, it can also leverage other word embeddings to evaluate the similarity between model-generated and reference text. The BERTScore of a single text pair is calculated according to the relationship:
   \begin{align}
    \text{BERTScore} = \frac{2 \times P \times R}{P + R},
   \end{align}
   where $P$ represents the ratio of the maximum cosine similarity score between tokens in the model-generated text and the reference text to the numbers of tokens in the model-generated text and $R$ represents the ratio of the maximum cosine similarity score between tokens in the model-generated text and the reference text to the numbers of tokens in the reference text. The BERTScore of the model is the average of BERTScores across all text pairs.

\paragraph{RadGraph F1} RadGraph F1 is a novel metric that measures overlap in clinical entities and relations extracted from radiology reports \cite{yu2023radgraph}. The RadGraph F1 score is computed in the following way. First, the RadGraph model maps model-generated and reference reports into graph representations with clinical entities represented as nodes and their relations as edges between them. Second, the number of nodes that match between the two graphs based on clinical entity text and labels (entity type) is determined. Third, the number of edges that match between the two graphs based on their start and end entities and labels (relation type) is calculated.  Lastly, the F1 score is separately computed for clinical entities and relations, and then the RadGraph F1 score for a report pair is the average of these two scores. The overall model performance is determined by averaging RadGraph F1 scores across all report pairs.
      
\paragraph{Human evaluation} Human evaluation plays a crucial role in assessing VLMS's quality in medical RG. Human evaluation can be performed for RG in various ways.  For instance, in \cite{jeong2023xrem}, expert radiologists evaluate the performance of the X-REM model in the RG task as follows. Initially, each report is segmented into lines, and radiologists assign scores to each line based on five error categories. These scores reflect the severity of errors, with higher values indicating more severe errors. Two metrics are utilized to obtain a comprehensive measure of the overall severity of errors in a report. Maximum Error Severity (MES) represents the highest score across all lines in the report. In contrast, Average Error Severity (AES) is calculated by averaging the scores across all lines in the report. According to radiologists, $18\%$ of model-generated reports received an MES score of $0$, while $24\%$ received an AES score of $0$.

\paragraph{Additional Evaluation Metrics for Report Generation}
The next few metrics are designed for classification evaluation, and RG can be viewed as such a task. In \cite{moon2022medvill}, \cite{lee2023unixgen}, and \cite{pellegrini2023radialog}, these metrics are computed based on the 14 labels obtained from applying the CheXpert \cite{irvin2019chexpert} or CheXbert \cite{smit2020chexbert} labeler to the reference reports as well as the model-generated reports. In this context, reports bearing accurate diagnosis labels are categorized as positive, while those with inaccurate labels are regarded as negative. The following metrics are also called clinical efficacy metrics.
\begin{itemize}

  \item \emph{Accuracy} measures the ratio of all positive predictions to the total number of predictions.

\item \emph{Precision} measures the accuracy of positive predictions made by a model. The precision score is calculated by considering the ratio of true positive predictions to the total number of instances that the model predicted as positive: 
  \begin{align}
  \text{Presicion} = \frac{\text{True Positivies}}{\text{True Positivies + False Positivies}}.
  \end{align} 
  High precision indicates that the model has a low rate of false positives.

\item \emph{Recall} is a metric that assesses the ability of a model to predict all positive classes. Recall is defined as the ratio of correctly predicted positive observations to the total actual positives: 
  \begin{align} 
  \text{Recall} = \frac{\text{True Positives}}{\text{True Positives + False Negatives}}.
  \end{align}
   The high recall means that the model effectively identifies most of the actual positive instances.

\item \emph{F1 Score} assesses the overall model's performance by balancing precision and recall into a single value. The F1 score is defined as:
   \begin{align}
   \text{F1}= \frac{2 \times \text{Precision} \times \text{\emph{Recall}}}{\text{Precision + False Recall}}.
   \end{align}
   The F1 scores range from 0 to 1, with higher values indicating better performance. In multi-class classification, it is common to compute the macro-F1 score by averaging the F1 scores calculated independently for each class. This method ensures an unbiased evaluation of the model's performance across all classes, assigning equal importance to each class, irrespective of its size or prevalence in the dataset.
\end{itemize}

\subsubsection{Evaluation Metrics for VQA}
The common benchmark datasets for medical VQA include VQA-RAD \cite{lau2018vqarad}, SLAKE \cite{liu2021slake}, and PathVQA \cite{he2020pathvqa}. For more information on these datasets, see \cref{subsec:datasets}. Various metrics are available for VQA evaluation, and many of those used for RG can also be applied to VQA. To avoid redundancy with already mentioned metrics, only a few are highlighted below.

\paragraph{Accuracy} Accuracy is a fundamental metric for gauging overall model correctness in VQA evaluation. It is determined by calculating the proportion of correctly predicted answers to the total number of questions. Sometimes, the average accuracy is computed by applying the model to various testing datasets, providing a comprehensive assessment of its performance across diverse scenarios. For a detailed comparison of accuracies among different medical VLMs discussed in \cref{subsec:models}, refer to \cref{tab:accuracies}.

\paragraph{Exact Match} Exact match metric computes the ratio of generated answers that match exactly (excluding punctuation) the correct answer. However, this measure is rather strict, as it may not give credit to valuable answers that, despite being semantically correct, diverge from an exact lexical match with the correct answer. This metric is more suitable for evaluating answers to close-ended questions than open-ended ones.
 
 \paragraph{Human Evaluation} Human evaluation is valuable for assessing a model's performance and applies not only to tasks such as VQA but also to RG. Human evaluation can be performed for VQA in various ways. For instance, in \cite{moor2023medflamingo}, the human evaluation process of MedFlamingo model employs an application featuring a user-friendly interface. Within this interface, medical experts are empowered to evaluate each VQA problem individually, assigning scores ranging from 0 to 10.  The final scores of the few-shot performance are $5.61$ on VQA-RAD, $1.81$ on PathVQA, and $4.33$ on the specifically curated Visual USMLE dataset. In contrast, the scores for zero-shot performance are lower, with $3.82$ on RAD-VQA, $1.72$ on PathVQA, and $4.18$ on Visual USMLE.  

\begin{table}[hp]
 \caption{A list of medical VLMs developed for VQA and RG.}
    \label{tab:vlms}
    \centering
    \scalebox{0.62}{\begin{tabular}{cccccccc}
    \hline
    \textbf{Model} & \textbf{Stream} & \textbf{Decoder} & \textbf{Architecture} & \textbf{VQA} & \textbf{RG} & \textbf{Datasets} & \textbf{Code}\\
    \hline
    \hline
   \multirow{2}{*}{\textbf{MedViLL}} & \multirow{3}{*}{single} & \multirow{3}{*}{No} & \multirow{3}{*}{RN50 + BERT} & \multirow{3}{*}{+} & \multirow{3}{*}{+} & \multirow{2}{*}{MIMIC-CXR,} & \multirow{3}{*}{\href{https://github.com/SuperSupermoon/MedViLL}{GH}}\\
    \multirow{2}{*}{\cite{moon2022medvill}} & & &  & & & \multirow{2}{*}{Open-I, VQA-RAD} & \\
     & & &  & & &  & \\
    \multirow{2}{*}{\textbf{PubMedCLIP}} & \multirow{3}{*}{dual} & \multirow{3}{*}{No} &  ViT-B/32 or RN50 or & \multirow{3}{*}{+} & \multirow{3}{*}{--} & \multirow{2}{*}{ROCO, SLAKE,} & \multirow{3}{*}{\href{https://github.com/sarahESL/PubMedCLIP/tree/main/PubMedCLIP}{GH}}\\
     \multirow{2}{*}{\cite{eslami2023pubmedclip}} & &  & RN50$\times$4  + Transformer & & & \multirow{2}{*}{VQA-RAD}  & \\
       & &  &  + BAN & & & & \\
    \multirow{2}{*}{\textbf{RepsNet}} & \multirow{3}{*}{dual} & \multirow{3}{*}{No} & \multirow{2}{*}{ResNeXt-101 + BERT} & \multirow{3}{*}{+} & \multirow{3}{*}{+} & \multirow{2}{*}{VQA-RAD,} & \multirow{3}{*}{\href{https://sites.google.com/view/repsnet}{Site}}\\
     \multirow{2}{*}{\cite{tanwani2022repsnet}} & & & \multirow{2}{*}{+ BAN + GPT-2} & & & \multirow{2}{*}{IU-Xray}  & \\
     & &  & & & & & \\
    \multirow{2}{*}{\textbf{BiomedCLIP}} & \multirow{3}{*}{dual} & \multirow{3}{*}{No} & ViT-B/16 & \multirow{3}{*}{+} & \multirow{3}{*}{--} & \multirow{2}{*}{PMC-15, SLAKE,} & \multirow{3}{*}{\href{https://huggingface.co/microsoft/BiomedCLIP-PubMedBERT_256-vit_base_patch16_224}{HF}}\\
    \multirow{2}{*}{\cite{zhang2023biomedclip}} & & & + PubMedBERT & & & \multirow{2}{*}{VQA-RAD} & \\
     & & & + METER & & & & \\
     \multirow{2}{*}{\textbf{UniXGen}} & \multirow{3}{*}{single} & \multirow{3}{*}{No} & \multirow{3}{*}{VQGAN + Transformer} & \multirow{3}{*}{--} & \multirow{3}{*}{+} & \multirow{3}{*}{MIMIC-CXR} & \multirow{3}{*}{\href{https://github.com/ttumyche/UniXGen}{GH}}\\
    \multirow{2}{*}{\cite{lee2023unixgen}} & & &  & & & & \\
     & & & & & & & \\
     & \multirow{5}{*}{dual} & \multirow{5}{*}{No} & Swiss Transformer & \multirow{5}{*}{+} & \multirow{5}{*}{--} & PMCPM, ROCO & \multirow{5}{*}{\href{https://github.com/GanjinZero/RAMM}{GH}}\\
    \multirow{2}{*}{\textbf{RAMM}} & & &  + PubMedBERT  & & & MIMIC-CXR, & \\
    \multirow{2}{*}{\cite{yuan2023ramm}} & & & + multimodal encoder w/ & & & SLAKE, VQA-RAD,  & \\
      & & &  retrieval-atten. module & & &  VQA-Med 2019, & \\
      & & &  & & & VQA-Med 2021 & \\
      & \multirow{4}{*}{dual} & \multirow{4}{*}{No} & \multirow{2}{*}{ALBEF} & \multirow{4}{*}{--} & \multirow{4}{*}{+} &  & \multirow{4}{*}{\href{https://github.com/rajpurkarlab/X-REM}{GH}}\\
    \textbf{X-REM} & & & \multirow{2}{*}{(ViT-B/16 + BERT}  & & & MIMIC-CXR, & \\
   \cite{jeong2023xrem} & & & \multirow{2}{*}{+ multimodal encoder)} & & & MedNLI, RadNLI & \\
     & &  & & & & & \\
    &  \multirow{5}{*}{single} &  \multirow{5}{*}{No}  & &  \multirow{5}{*}{+} & \multirow{5}{*}{--} &  ROCO; MedDialog,  
    &  \multirow{5}{*}{\href{https://github.com/cambridgeltl/visual-med-alpaca}{GH}}\\
    \textbf{Visual}   & & & DePlot or Med-GIT & &   & MEDIQA QA,    & \\
    \textbf{Med-Alpaca} & & & + prompt manager &  & &  MEDIQA RQE, & \\
    \cite{shu2023vismedalpaca} & &  & +LLaMa-7B & & & MedQA, PubMedQA & \\
       & & &  &  & & + GPT-3.5-Turbo & \\
      & \multirow{6}{*}{dual} & \multirow{6}{*}{No} &  ALBEF  & \multirow{6}{*}{--} & \multirow{6}{*}{+} & & \multirow{6}{*}{--}\\
      & & & + FAISS retriever  & &  &  & \\
     \textbf{CXR-RePaiR-Gen} & & & + prompt manager  & & & CXR-PRO, & \\
     \cite{ranjit2023cxrrepairgen}  & & &  + text-davinci-003 & & & MS-CXR & \\
      & & &  or GPT-3.5-Turbo &  & &  & \\
      & & &  or GPT-4 &  & &  & \\
    \multirow{2}{*}{\textbf{LLaVa-Med}} & \multirow{3}{*}{single} & \multirow{3}{*}{No} & \multirow{2}{*}{ViT-L/14 + projection}   & \multirow{3}{*}{+} & \multirow{3}{*}{--} & PMC-15 + GPT-4, & \multirow{3}{*}{\href{https://github.com/microsoft/LLaVA-Med}{GH}}\\
     \multirow{2}{*}{\cite{li2023llavamed}}  & & & \multirow{2}{*}{layer + LLaMa-7B}  & &  & VQA-RAD, SLAKE, & \\
        & & &  & & & PathVQA & \\
     & \multirow{4}{*}{single} & \multirow{4}{*}{No} &  \multirow{2}{*}{MedCLIP + linear}  & \multirow{4}{*}{+} & \multirow{4}{*}{+} & & \multirow{4}{*}{\href{https://github.com/mbzuai-oryx/XrayGPT}{GH}}\\
   \textbf{XrayGPT}  & & & \multirow{2}{*}{transformation  layer} &  &  & MIMIC-CXR\\
   \cite{thawkar2023xraygpt}  & & &  \multirow{2}{*}{+ Vicuna-7B}   &  &  & Open-I & \\
      & &  & & & & & \\
   \multirow{2}{*}{\textbf{CAT-ViL DeiT}} & \multirow{3}{*}{single} & \multirow{3}{*}{No} & RN18 + tokenizer & \multirow{3}{*}{+} & \multirow{3}{*}{--} & \multirow{2}{*}{EndoVis 2017,}  & \multirow{3}{*}{\href{https://github.com/longbai1006/CAT-ViL}{GH}} \\
    \multirow{2}{*}{\cite{bai2023catvil}}  &  &  & + CAT-ViL fusion  &  &  & \multirow{2}{*}{EndoVis 2018} & \\
      & &  & module + DeiT & & &   & \\
    & \multirow{5}{*}{dual} & \multirow{5}{*}{Yes}  & & \multirow{5}{*}{+} & \multirow{5}{*}{--} & \multirow{2}{*}{ROCO, MedICaT,} & \multirow{5}{*}{\href{https://github.com/pengfeiliHEU/MUMC}{GH}}\\
    \multirow{2}{*}{\textbf{MUMC}} & &  & ViT-B/12 + BERT   & & & \multirow{2}{*}{ImageCLEF Caption,} & \\
   \multirow{2}{*}{\cite{li2023mumc}} &  & & + multimodal encoder & & & \multirow{2}{*}{VQA-RAD, SLAKE}  &\\
    & & & + answer decoder & & & \multirow{2}{*}{PathVQA}  &\\
    & & &  &  & &  & \\
    \multirow{2}{*}{\textbf{Med-Flamingo}} & \multirow{3}{*}{single} & \multirow{3}{*}{No}  & \multirow{2}{*}{ViT-L/14 + perceiver}  & \multirow{3}{*}{+} & \multirow{3}{*}{--}& MTB, PMC-OA, & \multirow{3}{*}{\href{https://github.com/snap-stanford/med-flamingo}{GH}}\\
   \multirow{2}{*}{\cite{moor2023medflamingo}} & & & \multirow{2}{*}{resampler + LLaMa-7B} & & & VQA-RAD, PathVQA,  &\\
    &  & &  & & & Visual USMLE &\\
    & \multirow{4}{*}{single} & \multirow{4}{*}{No}  & \multirow{2}{*}{BioViL-T + BERT}  & \multirow{4}{*}{+} & \multirow{4}{*}{+}&   & \multirow{4}{*}{\href{https://github.com/ChantalMP/RaDialog}{GH}}\\
    \textbf{RaDialog} & &  & \multirow{2}{*}{+ prompt manager} & & & MIMIC-CXR, & \\
  \cite{pellegrini2023radialog} & & & \multirow{2}{*}{+ Vicuna-7B} & & &  Instruct & \\
   & & &  & & &  &\\
    \hline
\end{tabular}}
\end{table}

\subsection{Medical Models}\label{subsec:models}

In this part of the review paper, we provide an overview of existing medical VLMs tailored for VQA and/or RG.  The information is organized chronologically based on the first appearance of the model. Our focus is mainly on recently introduced open-source or publicly available models. A summary of these VLMs is presented in \cref{tab:vlms}.

\subsubsection{Medical Vision Language Learner (MedViLL)}
MedViLL can process medical images to generate associated reports \cite{moon2022medvill}. The model employs ResNet-50 \cite{he2016resnet}, trained on ImageNet \cite{deng2009imagenet}, for extracting visual features $v$. The model also leverages a base BERT \cite{devlin2019bert} embedding layers to extract textual features $t$ from clinical reports, which are initially segmented into a sequence of tokens using a WordPiece \cite{wu2016wordpiece} tokenizer. Both textual and visual features incorporate positional information to capture the spatial relationships and sequential order of elements in the input data. To generate a cross-modal representation, vectors $v$ and $t$, along with special tokens [CLS], [SEP]\textsubscript{V}, [SEP]\textsubscript{L}  are concatenated in a single vector as follows: $(CLS, v, SEP_V, t, SEP_L)$. The cross-modal representations are then fed into the BERT model. The MedViLL is pre-training on two tasks: MLM and ITM. The MLM task employs a bidirectional auto-regressive (BAR) self-attention mask, promoting the integration of image and language features. For MLM, a negative log-likelihood loss function is used. The ITM task encourages learning visual and textual features by predicting matching pairs and employs a loss function based on predictions for matching and non-matching pairs. The model is pre-trained on $89,395$ image-report pairs from the MIMIC-CXR \cite{johnson2019mimiccxr} dataset and then fine-tuned for downstream tasks on $3,547$ pairs from the Open-I \cite{demnerfushman2015openi} dataset, an additional dataset comprising radiographic image-report pairs. Only AP view X-rays are included in the analysis of both datasets. VQA is performed on the VQA-RAD \cite{lau2018vqarad} dataset (see \cref{tab:accuracies}), including open and close-ended questions, where the output representation of [CLS] is used to predict a one-hot encoded answer. For radiology RG fine-tuning, the model uses a sequence-to-sequence (S2S) mask instead of BAR and generates reports by sequentially recovering MASK tokens. RG is evaluated on MIMIC-CXR \cite{johnson2019mimiccxr} and Open-I \cite{demnerfushman2015openi}. MedViLL achieves a BLEU-4 score of $0.066$, a perplexity value of $4.185$, and using a CheXpert labeler \cite{irvin2019chexpert} an accuracy of $84.1\%$, a precision value of $0.698$, a recall value of $0.559$, and an F1 score of $0.621$ on MIMIC-CXR. Additionally, it achieves a BLEU-4 score of $0.049$, a perplexity value of $5.637$, an accuracy of $73.4\%$, a precision value of $0.512$, a recall value of $0.594$, and an F1 score of $0.550$ on Open-I.

\subsubsection{PubMedCLIP}
PubMedCLIP \cite{eslami2023pubmedclip} is a CLIP-based \cite{radford2021clip} model pre-trained on ROCO \cite{pelka2018roco} dataset, consisting of over 80K image-caption pairs sourced from PMC articles. The model utilizes a CLIP text encoder, which is based on the Transformer \cite{vaswani2017attention} architecture, and three distinct CLIP visual encoders: ViT-B/32 \cite{dosovitskiy2021vit}, ResNet-50, and  ResNet-50$\times$4 \cite{he2016resnet}. Following the contrastive learning approach in CLIP, the model generates joint representations by computing cosine similarity between textual and visual features. The pre-training objective involves the computation of \emph{cross-entropy loss} values for both vision and language. These losses are then averaged to derive an overall loss value. Following pre-training, the model is repurposed as a pre-trained visual encoder for VQA. The visual feature in VQA is the concatenation of the model's output with a convolutional denoising autoencoder (CDAE) \cite{masci2011cdae} output, an image denoising module. The question is encoded using a GloVe \cite{pennington2014glove} word embedding followed by an LSTM \cite{hochreiter1997lstm}. The image and question features are combined using bilinear attention networks (BAN) \cite{kim2018ban}, and the resulting representations are passed through an answer classifier, which is a two-layer feedforward NN. The VQA loss is determined by combining the classification and image reconstruction losses. During the VQA fine-tuning, the SLAKE (English) \cite{liu2021slake} and VQA-RAD \cite{lau2018vqarad} datasets, comprising both open- and close-ended questions, are employed. The model's effectiveness is evaluated in the context of two existing Medical VQA (MedVQA)  methods: Mixture of Enhanced Visual Features (MEVF) \cite{zhan2020mevf} and question-conditioned reasoning (QCR) \cite{liu2023condreason}. The assessment involved replacing the visual encoder component in MEVF and QCR with PubMedCLIP and subsequently evaluating the model's performance. PubMedCLIP in the QCR framework achieves better accuracies on VQA-RAD and SLAKE datasets than in the MEVF framework. The highest accuracies of PubMedCLIP in the QCR framework on both datasets are shown in \cref{tab:accuracies}.

\subsubsection{RepsNet}
RepsNet is designed for VQA tasks. It can generate automated medical reports and interpret medical images. The model employs a modified version of the pre-trained ResNeXt-101 \cite{xie2016resnext} as its image encoder and utilizes pre-trained BERT \cite{devlin2019bert} as the text encoder, with text tokenization done through WordPiece \cite{wu2016wordpiece}. Fusion of image and question features is achieved using BAN \cite{kim2018ban}. To align images with textual descriptions, the model employs bidirectional contrastive learning \cite{chen2020contlearningvisrepr}. For VQA tasks, the model is fine-tuned and evaluated on VQA-RAD \cite{lau2018vqarad} (see \cref{tab:accuracies}). In contrast, for RG, fine-tuning and evaluation are done using IU-Xray \cite{demnerfushman2015openi} dataset. The model categorizes answers through classification for close-ended questions and generates answers using the modified version of GPT-2 language decoder based on image features and prior context. The BLEU-2 and BLEU-4 scores of RepsNet on the IU-Xray dataset are $0.44$ and $0.27$, respectively.
    
\subsubsection{BiomedCLIP}
BiomedCLIP is pre-trained on the specifically curated PMC-15 dataset that consists of $15$ M figure-caption pairs derived from the PMC articles \cite{zhang2023biomedclip}. However, the models is not publicly available. The model architecture is similar to CLIP \cite{radford2021clip}, except that the text encoder is a pre-trained PubMedBERT \cite{gu2021pubmedbert} model with WordPiece tokenizer \cite{wu2016wordpiece}. The model uses ViT-B/16 \cite{dosovitskiy2021vit} as the visual data encoder. During pre-training, the model adopts a contrastive learning approach, and to mitigate memory usage, it utilizes the sharding contrastive loss \cite{cherti2022shconrtloss}. For adaptation to VQA, the model incorporates the METER \cite{dou2022meter} framework. This involves deploying a Transformer-based co-attention multimodal fusion module that produces cross-modal representations. These representations are then fed into a classifier for the final prediction of answers. The model is evaluated on VQA-RAD \cite{lau2018vqarad} and SLAKE (English) \cite{liu2021slake} datasets (see \cref{tab:accuracies}).

\subsubsection{Unified chest X-ray and report Generation model (UniXGen)}
UniXGen is a unified model that can generate both reports and view-specific X-rays \cite{lee2023unixgen}. The model tokenizes chest X-rays leveraging VQGAN \cite{esser2021vqgan}, a generative model that amalgamates generative adversarial networks (GANs) with vector quantization (VQ) techniques. VQGAN employs an encoder to transform input images into continuous representations, subsequently using vector quantization to discretize them into learnable codebook vectors. Additionally, VQGAN incorporates a decoder, translating these discrete codes back into images during the generation process. For chest X-rays, multiple views from the same study are tokenized into sequences of discrete visual tokens, demarcated by special tokens to distinguish perspectives. In the case of radiology reports, the model uses the byte-level BPE \cite{wang2020bbpe} tokenizer, augmented with sinusoid positional embedding for enhanced representation. The model is based on the Transformer architecture \cite{vaswani2017attention} with a multimodal causal attention mask, ensuring that each position in the sequence attends to all previous positions and not future ones. During training, multiple views of chest X-rays and a report embedding are concatenated randomly and fed into the Transformer. The model is optimized using the negative log-likelihood loss function. The model is trained on $208,534$ studies sampled from the MIMIC-CXR \cite{johnson2019mimiccxr} dataset. Each study contains at most three chest X-rays representing PA (from back to front), AP (from front to back), and lateral views. UniXGen achieves a BLEU-4 score of $0.050$, and using a CheXpert labeler \cite{irvin2019chexpert} a precision score of $0.431$, a recall value of $0.410$, and an F1 score of $0.420$ on MIMIC-CXR dataset.

\subsubsection{Retrieval-Augmented bioMedical Multi-modal Pretrain-and-Finetune Paradigm (RAMM)}
RAMM is a retrieval-augmented VLM tailored for biomedical VQA \cite{yuan2023ramm}. The model uses Swin Transformer \cite{liu2021swintransformer} as the image encoder and PubMedBERT \cite{gu2021pubmedbert} as the text encoder. The visual and textual features are then fused by the multimodal encoder, a 6-layer Transformer \cite{vaswani2017attention}. The model is pre-trained on the MIMIC-CXR \cite{johnson2019mimiccxr} and ROCO \cite{pelka2018roco} datasets along with a newly curated PMC-Patients-Multi-modal (PMCPM) dataset, consisting of $398,000$ image-text pairs sampled from PMC-OA \cite{lin2023pmcclip} dataset. The pre-training objective function of the model is the sum of three tasks: contrastive learning, ITM, and MLM. Using contrastive learning, the model aligns images and texts using the cosine similarity metric. The VQA task is viewed as a classification problem, and the model is optimized using the cross-entropy loss function. During model fine-tuning, the retrieval-attention module fuses the representations of the image-question input with four representations of the retrieved image-text pairs from the pre-trained datasets. This allows the model to focus on relevant parts of the retrieved information when generating answers. The model is evaluated on VQA-Med 2019 \cite{abacha2019vqamed19}, VQA-Med 2021 \cite{ionescu2021vqamed21}, VQA-RAD \cite{lau2018vqarad}, and SLAKE \cite{liu2021slake} datasets (see \cref{tab:accuracies}). 

 \subsubsection{Contrastive X-Ray REport Match (X-REM)}
 X-REM is a retrieval-based radiology RG model that uses an ITM score to measure the similarity of a chest X-ray image and radiology report for report retrieval \cite{jeong2023xrem}. The VLM backbone of the model is ALBEF \cite{li2021albef}. ALBEF utilizes ViT-B/16 \cite{dosovitskiy2021vit} as its image encoder and initializes the text encoder with the first 6 layers of the BERT \cite{devlin2019bert} base model. The multimodal encoder in ALBEF, responsible for combining visual and textual features to generate ITM scores, is initialized using the final six layers of the BERT base model. X-REM leverages ALBEF's pre-trained weights and performs further pre-training on X-rays paired with extracted impression sections ($2,192$ pairs), findings sections ($1,597$ pairs), or both ($2,192$ pairs) from the MIMIC-CXR \cite{johnson2019mimiccxr} dataset. Subsequently, the model is fine-tuned on the ITM task, where the scoring mechanism involves using the logit value for the positive class as the similarity score for image-text pairs. To address the positive skewness in medical datasets, 14 clinical labels obtained from the CheXbert \cite{smit2020chexbert} labeler are utilized. The model efficiently manages the computational burden associated with ITM scores by employing ALBEF's pre-aligned unimodal embeddings. This involves narrowing down the candidate reports based on high cosine similarity with the input image before computing ITM scores. Additionally, the text encoder undergoes fine-tuning on natural language inference (NLI) task, utilizing datasets such as MedNLI \cite{romanov2018mednli} and RadNLI \cite{miura2021radnli}. This step is crucial for preventing the retrieval of multiple reports with overlapping or conflicting information. X-REM achieves a BLEU-2 score of $0.186$ on the MIMIC-CXR (Findings only) dataset.  The BERTScore of the model is $0.386$ on MIMIC-CXR (Findings only) and is $0.287$ on MIMIC-CXR (Impressions and Findings). The human evaluation of X-REM is described in \cref{subsec:metrics}.
    
\subsubsection{Visual Med-Alpaca} 
Visual Med-Alpaca is a biomedical foundational model designed for addressing multimodal biomedical tasks like VQA \cite{shu2023vismedalpaca}. The model is constructed in the following way. First, image inputs go through a classifier to determine the appropriate module for transforming visual information into an intermediate text format. The currently supported modules include DePlot \cite{liu2023deplot}, utilized for interpreting plots and charts, and Med-GIT \cite{wang2022git}, fine-tuned specifically on the ROCO \cite{pelka2018roco}  dataset for understanding radiology images. The prompt manager then amalgamates textual information extracted from images and text inputs to construct the prompt for the LLM model, LLaMA-7B \cite{touvron2023llama}. However, before generating responses, LLaMa-7B undergoes both standard fine-tuning and LoRA \cite{hu2022lora} fine-tuning on a carefully curated set of $54,000$ medical question-answer pairs. The questions within this set are derived from question-answering datasets such as MEDIQA QA \cite{benabacha2019mediqa}, MEDIQA RQE \cite{benabacha2019mediqa}, MedQA \cite{jin2021medqa}, MedDialog \cite{zeng2020meddialog}, and PubMedQA \cite{jin2019pubmedqa}, with their corresponding answers synthesized using GPT-3.5-Turbo in the \emph{self-instruct} \cite{wang2023selfinstruct} manner. Human experts then meticulously filter and edit the obtained question-answer pairs to ensure quality and relevance. The evaluation of this model is still ongoing \cite{shu2023vismedalpaca}. 

 \subsubsection{Contrastive X-ray-Report Pair Retrieval based Generation (CXR-RePaiR-Gen)} 
 CXR-RePaiR-Gen is designed for radiology RG that incorporates the RAG framework to mitigate the issue of \emph{hallucinated references} \cite{ranjit2023cxrrepairgen}. The model leverages the pre-trained ALBEF  \cite{lan2019albert} previously utilized in CXR-ReDonE \cite{ramesh2022cxrredone}. The ALBEF model consists of a ViT-B/16 \cite{dosovitskiy2021vit} image encoder and the first 6 layers of BERT \cite{devlin2019bert} as the text encoder, producing contrastively aligned image and text embeddings. Textual features are indexed in a vector database, Facebook AI Similarity Search (FAISS). When given a radiology image input, embeddings from the reports or sentences corpus with the highest dot-product similarity to the image embedding are retrieved. The CXR-PRO \cite{ramesh2022cxrredone} dataset is employed for text retrieval to gather relevant impressions for generating the radiology report. The retrieved impression sections from the CXR-PRO dataset serve as the context for the prompt to an LLM, along with instructions to generate the radiology report. Two distinct prompts are employed for generating free-text reports: one for the text-davinci-003 model and another for RG in a conversational setting with the GPT-3.5-Turbo and GPT-4 models. The model is evaluated on MS-CXR \cite{boecking2022mscxr} and CXR-PRO datasets. There is no code provided for this model yet. CXR-RePaiR-Gen reaches a BERTScore score of $0.2865$ on the CXR-PRO dataset when based on GPT-4. Additionally, CXR-RePaiR-Gen achieves a score of $0.1970$ on MS-CXR when based on text-davinci-003. The model attains a RadGraph F1 score of $0.1061$ on the CXR-PRO dataset when based on GPT-4 and  $0.0617$ on the MS-CXR dataset when it is based on text-davinci-003. In these instances, the CXR-RePaiR-Gen utilizes three retrieval samples per input during the RAG process. 
    
\subsubsection{Large Language and Vision Assistant for BioMedicine (LLaVa-Med)} LLaVa-Med is an adaptation of the VLM LLaVa \cite{liu2023llava}, specifically tailored for the medical domain through training on instruction-following datasets \cite{li2023llavamed}.  The visual features are generated by the pre-trainedCLIP \cite{radford2021clip} visual encoder ViT-L/14 \cite{dosovitskiy2021vit}. The encoder can be substituted with BiomedCLIP \cite{zhang2023biomedclip}. These features are passed through a linear projection layer, which converts them into tokens, and then, together with the tokenized instructions, are fed into the LLM LLaMa-7B \cite{touvron2023llama}. The LLM can be substituted with Vicuna \cite{chiang2023vicuna}. After initializing with the general-domain LLaVA, the model undergoes fine-tuning using curriculum learning. First, the model tries to understand and connect visual elements in biomedical images to the corresponding words or descriptions in the language model's knowledge. To achieve that, a dataset consisting of $600,000$ image-caption pairs from the PMC-15 dataset, which was originally employed in the training of BiomedCLIP, is utilized. These image-caption pairs are transformed into an instruction-following dataset, where the instructions prompt the model to describe the corresponding image concisely or in detail.
Given the language instruction and image input, the model is then prompted to predict the original caption. During this stage, the visual encoder and language model weights are kept frozen, with updates exclusively applied to the linear projection layer. The second stage of training focuses on aligning the model to follow diverse instructions.  For this purpose, another instruction-following dataset is generated from PMC-15. For this dataset, instructions are designed to guide the GPT-4 model to generate multi-round questions and answers from the image caption and sentences from the original PMC paper that mentions the image \cite{li2023llavamed}. In this training phase, the model undergoes training on a set of $60,000$ images, each accompanied by its respective caption and multi-round questions and answers. Throughout this process, the weights of the visual encoder remain unchanged, preserving the previously acquired visual features. Meanwhile, the pre-trained weights of both the projection layer and the language model undergo continuous updates. This approach enables the model to effectively respond to a variety of instructions and perform well in generating dynamic and informative multi-round conversational content. Lastly, for VQA, the model is fine-tuned and evaluated on VQA-RAD \cite{lau2018vqarad}, SLAKE \cite{liu2021slake}, and PathVQA \cite{he2020pathvqa} (see \cref{tab:accuracies}).

\subsubsection{XrayGPT}
XrayGPT is a conversational medical VLM specifically developed for analyzing chest radiographs \cite{thawkar2023xraygpt}. The VLM uses MedCLIP \cite{wang2022medclip} as a vision encoder to generate visual features. These features undergo a meticulous transformation process: initially, they are mapped to a lower-dimensional space through a linear projection head and subsequently translated into tokens via a linear transformation layer. At its core, the model incorporates two text queries: (1) the assistant query plays a role in contextualizing the model's behavior and defining its purpose as ``You are a helpful healthcare virtual assistant'', (2) the doctor's query serves as a prompt that guides the model in providing information relevant to chest X-ray analysis. Tokens generated from a visual input are concatenated with the tokenized queries and then fed into the medical LLM,  which generates the summary of the chest x-ray. The LLM employed in this architecture is Vicuna-7B \cite{chiang2023vicuna}, fine-tuned on a rich dataset consisting of $100,000$ real conversations between patients and doctors, along with $20,000$ radiology conversations sourced from \url{ShareGPT.com}. During training, the weights of both the vision encoder and the LLM remain frozen while the weights in the linear transformation layer undergo updates. The model is first trained on $213,514$ image-text pairs from pre-processed MIMIC-CXR \cite{johnson2019mimiccxr} dataset and then on $3,000$ image-text pairs from Open-I \cite{demnerfushman2015openi} dataset. XrayGPT achieves $\text{ROUGE-1} = 0.3213$, $\text{ROUGE-2} = 0.0912$, and $\text{ROUGE-L} = 0.1997$ on MIMIC-CXR  dataset. 

\subsubsection{Co-Attention gaTed Vision-Language Data-efficient image Transformer (CAT-ViL DeiT)}
CAT-ViL DeiT stands out as a specialized VLM tailored for VQA within surgical scenarios, with a unique focus on answer localization \cite{bai2023catvil}. The architecture incorporates a ResNet-18 \cite{he2016resnet} as the visual encoder, pre-trained on ImageNet \cite{deng2009imagenet}, and a customized pre-trained BERT tokenizer \cite{seenivasan2022surgicalvqa} for the text encoder. Central to its functionality is the \emph{Co-Attention gaTed Vision-Language} (CAT-ViL) module, which enables interaction between visual and textual features and fuses them via a gating mechanism to obtain optimized multimodal embeddings. These features are then fused using a gating mechanism, yielding optimized multimodal embeddings. The model further integrates a pre-trained \emph{Data-efficient image Transformer} (DeiT) \cite{touvron2021deit} module to process these multimodal embeddings, aiming to acquire an optimal joint representation for comprehensive visual and textual understanding. In the context of VQA, the model adopts a standard classification head, while for answer localization within images, it employs the \emph{detection with transformers} (DETR) \cite{carion2020detr} head. The overall loss function comprises cross-entropy as the classification loss and L1-norm, along with the \emph{generalized intersection over union} (GIoU) \cite{rezatofighi2019giou}, serving as the localization loss. The model is trained on $1,560$ frames and $9,014$ QA pairs from the surgical datasets EndoVis 2018 \cite{allan2020endovis18}. The model achieved an accuracy of $61.92\%$ on the remaining data from EndoVis 2018 and $45.55\%$ on EndoVis 2017 \cite{allan2019endovis17} dataset.

\subsubsection{Masked image and text modeling with Unimodal and Multimodal Contrastive losses (MUMC)}
MUMC utilizes a ViT-B/12 \cite{dosovitskiy2021vit} as its image encoder, the first 6 layers of BERT \cite{devlin2019bert} as its text encoder, and the last 6 layers of BERT as its multimodal encoder \cite{li2023mumc}. The multimodal encoder incorporates cross-attention layers to align visual and textual features. For pre-training, the model employs a combination of contrastive learning, MLM, and ITM objectives. Also, the model utilizes a newly introduced \emph{masked image strategy}, randomly masking 25\% of image patches as a data augmentation technique. This exposes the model to a greater variety of visual contexts and enables learning representations that are more robust to partially occluded inputs. The pre-training is performed on the ROCO \cite{radford2021clip}, MedICaT \cite{subramanian2020medicat}, and Image Retrieval in Cross-Language Evaluation Forum (ImageCLEF)  caption \cite{ruckert2022imageclef} datasets. For downstream VQA tasks, an answering decoder is added on top of the multimodal encoder to generate answer text tokens. The encoder weights are initialized from pre-training, and the model is fine-tuned and evaluated on VQA-RAD \cite{lau2018vqarad}, SLAKE \cite{liu2021slake}, and PathVQA \cite{he2020pathvqa} (see \cref{tab:accuracies}).
    
\subsubsection{Med-Flamingo}
Med-Flamingo is a multimodal few-shot learner model based on the Flamingo \cite{alayrac2022flamingo} architecture, adapted to the medical domain \cite{moor2023medflamingo}. The model is pre-trained on the MTB \cite{moor2023medflamingo} dataset, a newly curated collection comprising $4,721$ segments from various Medical TextBooks, encompassing both textual content and images. Each segment is designed to contain at least one image and up to $10$ images, with a specified maximum length. Also, it is pre-trained on $1.3$ M image-caption pairs from the PMC-OA \cite{lin2023pmcclip} dataset. The Model's few-shot capabilities are achieved through training on these mixed text and image datasets, enabling it to generalize and perform diverse multimodal tasks with only a few examples. The model utilizes a pre-trained frozen CLIP vision encoder ViT-L/14 for visual feature generation. To convert these visual features into a fixed number of tokens, the model employs a module known as the \emph{perceiver resampler}, which is trained from scratch. Subsequently, these tokens, along with tokenized text inputs, undergo further processing in a pre-trained frozen LLM LLaMA-7B \cite{touvron2023llama}, enhanced with strategically inserted gated cross-attention layers that are also trained from scratch. This augmentation not only facilitates the learning of novel relationships but also bolsters training stability. The model's performance is evaluated on established benchmarks such as VQA-RAD \cite{lau2018vqarad} and PathVQA \cite{he2020pathvqa}, demonstrating its effectiveness in medical visual question-answering. The exact match scores for MedFlamingo demonstrate a few-shot performance of $0.200$ on VQA-RAD and $0.303$ on PathVQA. In contrast, the zero-shot performance yields an exact match score of $0.000$ on VQA-RAD and $0.120$ on PathVQA. Additionally, it is evaluated on a specifically created Visual United States Medical Licensing Examination (USMLE) dataset, comprising $618$ challenging open-ended USMLE-style questions augmented with images, case vignettes, and tables of laboratory measurements, covering a diverse range of medical specialties.  The human evaluation of the Med-Flamingo model on VQA-RAD, PathVQA, and Visual USMLE datasets is described in \cref{subsec:metrics}.

\subsubsection{RaDialog}
RaDialog is a VLM that integrates automated radiology RG with conversational assistance \cite{pellegrini2023radialog}. The model incorporates BioViL-T \cite{bannur2023biovilt}, a hybrid model that fuses the strengths of ResNet-50 \cite{he2016resnet} and Transformer \cite{vaswani2017attention} architectures. Pre-trained on radiology images and reports, BioViL-T serves as a vision encoder that generates patch-wise visual features. The extracted features undergo alignment through a BERT \cite{devlin2019bert} model, transforming them into a concise representation of $32$ tokens. The model incorporates the CheXpert classifier to offer organized findings in medical images. These findings are generated based on labels obtained from the CheXbert \cite{smit2020chexbert} model. The classifier is trained independently using labels predicted by CheXbert from the findings section of radiology reports. The model integrates visual features, structured findings, and the directive ``Write a radiology report'' into a singular prompt, which is used as input for the LLM, a Vicuna-7B \cite{chiang2023vicuna} model fine-tuned using LoRA \cite{hu2022lora}. The training is performed on X-ray image-report pairs from MIMIC-CXR \cite{johnson2019mimiccxr} dataset. RaDialog achieves a BLEU-4 score of $0.095$, ROUGE-L score of $0.2710$, METEOR score of $0.14$, and BERTScore of $0.400$ on the MIMIC-CXR dataset. To address the challenge of catastrophic forgetting during training and ensure the model's capability across diverse downstream tasks, it is specifically trained on the newly created Instruct \cite{pellegrini2023radialog} dataset.  This dataset is meticulously curated to encompass a spectrum of 8 diverse tasks: RG, NLE, complete CheXpert QA, binary CheXpert QA, region QA, summarization, report correction, and reformulation report using simple language. Carefully formulated prompts accompany each task, tailored to elicit specific responses from the model. For instance, some prompts involve answering questions about particular X-ray regions. RaDialog trained on the Instruct dataset achieves an F1 score of $0.397$ on the binary CheXpert QA task and $0.403$ on the complete CheXpert QA task. In contrast, RaDialog without being trained on Instruct achieves lower F1 scores of $0.018$ and $0.098$, respectively.

\begin{table}[t]
    \caption{The comparison of medical VLMs' accuracies on VQA tasks. The underlined accuracies are the highest for a specific dataset.}
    \label{tab:accuracies}
    \centering
    \scalebox{0.62}{\begin{tabular}{ccccccccc}
    \hline
    \multirow{4}{*}{\textbf{Model}} & \multirow{2}{*}{\textbf{SLAKE}} & \multirow{2}{*}{\textbf{SLAKE}}  & \multirow{2}{*}{\textbf{VQA-RAD}} & \multirow{2}{*}{\textbf{VQA-RAD}} & \multirow{2}{*}{\textbf{PathVQA}} & \multirow{2}{*}{\textbf{PathVQA}} &  & \\ 
     & \multirow{2}{*}{\textbf{open}} & \multirow{2}{*}{\textbf{close}} & \multirow{2}{*}{\textbf{open}} & \multirow{2}{*}{\textbf{close}}  & \multirow{2}{*}{\textbf{open}} & \multirow{2}{*}{\textbf{close}} & \textbf{VQA-Med}& \textbf{VQA-Med}\\ 
    & \multirow{2}{*}{\textbf{-ended}} & \multirow{2}{*}{\textbf{-ended}} & \multirow{2}{*}{\textbf{-ended}} & \multirow{2}{*}{\textbf{-ended}}  & \multirow{2}{*}{\textbf{-ended}} &  \multirow{2}{*}{\textbf{-ended}} & \textbf{2019} & \textbf{2021} \\ 
    & & & & & & & &  \\
    \hline
     & & & & & & & &  \\
    \textbf{MedViLL} & \multirow{2}{*}{--} & \multirow{2}{*}{--} & \multirow{2}{*}{59.50\%} & \multirow{2}{*}{77.70\%} & \multirow{2}{*}{--} & \multirow{2}{*}{--} & \multirow{2}{*}{--} & \multirow{2}{*}{--}\\
    \cite{moon2022medvill} & & & & & & & &  \\
     & & & & & & & &  \\
    \textbf{PubMedCLIP} & \multirow{2}{*}{78.40\%} & \multirow{2}{*}{82.50\%} & \multirow{2}{*}{60.10\%} & \multirow{2}{*}{80.00\%} & \multirow{2}{*}{--} & \multirow{2}{*}{--} & \multirow{2}{*}{--} & \multirow{2}{*}{--}\\
    \cite{eslami2023pubmedclip} & & & & & & & & \\
     & & & & & & & &  \\
    \textbf{RepsNet} & \multirow{2}{*}{--} & \multirow{2}{*}{--} & \multirow{2}{*}{--} & \multirow{2}{*}{\underline{87.05\%}} & \multirow{2}{*}{--} & \multirow{2}{*}{--} & \multirow{2}{*}{--} & \multirow{2}{*}{--}\\
    \cite{tanwani2022repsnet} & & & & & & & & \\
     & & & & & & & &  \\
    \textbf{BioMedCLIP} & \multirow{2}{*}{\underline{82.50\%}} & \multirow{2}{*}{89.70\%} & \multirow{2}{*}{67.60\%} & \multirow{2}{*}{79.80\%} & \multirow{2}{*}{--} & \multirow{2}{*}{--} & \multirow{2}{*}{--} & \multirow{2}{*}{--}\\
    \cite{zhang2023biomedclip} & & & & & & & & \\
     & & & & & & & &  \\
    \textbf{RAMM} & \multirow{2}{*}{82.48\%} & \multirow{2}{*}{\underline{91.59\%}} & \multirow{2}{*}{67.60\%} & \multirow{2}{*}{85.29\%} & \multirow{2}{*}{--} & \multirow{2}{*}{--} & \multirow{2}{*}{\underline{82.13\%}} & \multirow{2}{*}{\underline{39.20\%}}\\
    \cite{yuan2023ramm} & & & & & & & &  \\
     & & & & & & & &  \\
    \textbf{LLaVa-Med} & \multirow{2}{*}{--} & \multirow{2}{*}{84.19\%} & \multirow{2}{*}{--} & \multirow{2}{*}{85.34\%} & \multirow{2}{*}{--} & \multirow{2}{*}{ \underline{91.21\%}} & \multirow{2}{*}{--} & \multirow{2}{*}{--}\\
    \cite{li2023llavamed} & & & & & & & & \\
     & & & & & & & &  \\
    \textbf{MUMC} & \multirow{2}{*}{--} & \multirow{2}{*}{--} & \multirow{2}{*}{\underline{71.50\%}} & \multirow{2}{*}{84.20\%} & \multirow{2}{*}{\underline{39.00\%}} & \multirow{2}{*}{90.4\%} & \multirow{2}{*}{--} & \multirow{2}{*}{--}\\
    \cite{li2023mumc} & & & & & & & & \\
     & & & & & & & &  \\
    \hline
    \end{tabular}}
\end{table}

\section{Challenges and Potential Future Directions}\label{sec:future_work}
In medical AI, the future holds great promise and, concurrently, poses notable challenges \cite{acosta2022biomedicalai}. As technology advances, the integration of VLMs into the healthcare sector has the potential to revolutionize diagnostics, treatment planning, and patient care. Future medical VLMs may offer enhanced capabilities in understanding complex clinical scenarios, generating detailed reports of medical images, and facilitating seamless communication between healthcare professionals and AI systems. However, these advancements come with challenges. 

A significant challenge in developing effective medical VLMs is the limited availability of ML-ready diverse and representative medical datasets. This limitation restricts the comprehensive training of VLMs, impeding their ability to understand the complexities of diverse and rare clinical scenarios \cite{moor2023medflamingo}. VLMs with large context windows and RAG present a potential solution by increasing the model's context through the incorporation of retrieved relevant information. While RAG usually involves a frozen model during training, exploring the pre-training of VLMs within the RAG framework opens up a new avenue of research \cite{zhao2023multimodalrag}. This innovative approach could potentially enhance the robustness of VLMs, especially in handling new and unforeseen medical cases. Furthermore, the pressing concerns surrounding patient data privacy highlight the need for innovative solutions, like federated learning (FL). FL offers a promising strategy to alleviate the scarcity of medical data while prioritizing patient privacy \cite{zhang2021fedlearning}.  In this decentralized learning method, models are trained across multiple institutions, and only model weights are shared, not the data. Thus, it effectively addresses major concerns about patient privacy while enabling collaborative model training across diverse datasets.

Traditional metrics may fall short in capturing the nuanced complexities of clinical language, posing a barrier to reliable evaluations of VLM performance \cite{yu2023radgraph}. This issue becomes particularly evident when evaluating the accuracy of medical reports or addressing open-ended medical queries, where metrics need to discern clinically relevant distinctions. Therefore, the development and adoption of specialized metrics tailored for medical RG and VQA is imperative. Such metrics are pivotal not only for evaluating model performance but also for assessing aspects like generalization, efficiency, and robustness. Establishing these metrics will significantly contribute to fostering precise evaluations and continual advancements in the capabilities of medical VLMs.

The issue of hallucinations in generative VLMs poses a significant challenge to their reliability and practical application \cite{liu2024vlmhallucination}. Hallucinations refer to instances where VLMs generate outputs that are not grounded in the provided images or are inconsistent with the established knowledge. In medical contexts, these hallucinations can have serious consequences, leading to inaccurate diagnostic information or treatment recommendations. One identified root cause of hallucinations is the lack of alignment between visual and textual information \cite{sun2023llavarlhf}. Training VLMs to effectively align these data modalities is crucial in mitigating the risk of hallucinations. For instance, LLaVA-RLHF \cite{sun2023llavarlhf} achieved hallucination reduction by incorporating RLHF to align different modalities. Further research is needed into building medical VLMs that base their generation on the factual medical knowledge of minimal hallucination.

Overcoming catastrophic forgetting poses an additional challenge in the development of medical VLMs. Catastrophic forgetting occurs when a model learning new information inadvertently erases or distorts previously acquired knowledge, potentially compromising its overall competence. Striking a balance during fine-tuning can be crucial; moderate fine-tuning can be helpful to adapt the model to a specific task, while excessive fine-tuning can lead to catastrophic forgetting \cite{zhai2023catasforgettingfinetun, khan2023importance}. Leveraging methodologies from continual learning \cite{wang2023contlearning, zhou2023proof, cai2024dynamictransformer, khan2023importance, khan2024brain} might be useful in the context of medical VLMs, where the ability to adapt and accumulate knowledge across diverse clinical tasks is paramount. Continual learning focuses on training models to sequentially learn from and adapt to new data over time while retaining knowledge from previously encountered tasks \cite{khan2024brain}. Also, incorporating adapters within the framework of continual learning can be a valuable tool in mitigating catastrophic forgetting \cite{zhang2023adapterincontlearning}.

Finally, clinical validation and adoption of VLMs necessitate a collaborative bridge between medical experts and AI/ML researchers. Trust, alignment with clinical needs, and ethical deployment are critical components for successfully integrating these models into healthcare workflows. Establishing robust collaborations ensures a dynamic synergy, combining domain expertise with technological advancements. This synergy is essential for the responsible and effective deployment of medical VLMs in healthcare.

\section*{Declaration of Competing Interest}
The authors declare that they have no known competing financial interests or personal relationships that could have appeared to influence the work reported in this paper.

\section*{Acknowledgements}
This work was partly supported by NSF awards 1903466, 2234836, and 2234468.



\printbibliography[title={References}]



\end{document}